\documentclass{article}

\usepackage{microtype}
\usepackage{graphicx}
\usepackage{subfigure}
\usepackage{booktabs} 

\usepackage{hyperref}

\usepackage[accepted]{icml2025}

\usepackage{amsmath}
\usepackage{amssymb}
\usepackage{mathtools}
\usepackage{amsthm}
\usepackage{multirow}
\usepackage{pgfplots}
\usepackage{subcaption} 
\usepackage{pifont}
\usepackage{float}
\usepackage{marvosym}

\newcommand{\Checkmark}{\ding{51}}       
\newcommand{\XSolidBrush}{\ding{55}}    
\newcommand{\yx}[1]{{\color{blue} #1}}

\usepackage[capitalize,noabbrev]{cleveref}

\theoremstyle{plain}

\theoremstyle{definition}

\theoremstyle{remark}

\usepackage[textsize=tiny]{todonotes}

\icmltitlerunning{IPSeg: Image Posterior Mitigates Semantic Drift in Class-Incremental Segmentation}

\begin{document}

\twocolumn[{
\icmltitle{IPSeg: Image Posterior Mitigates Semantic Drift in \\Class-Incremental Segmentation}



\icmlsetsymbol{equal}{*}
\begin{center}
\textbf{
Xiao Yu$^{1,2,\ast}$ \quad
Yan Fang$^{1,2,\ast}$ \quad
Yao Zhao$^{1,2}$ \quad
Yunchao Wei$^{1,2,}$\textsuperscript{\Letter} \quad
}\\
{
$^1$ Institute of Information Science, Beijing Jiaotong University \quad
$^2$ Visual Intelligence + X International Joint Laboratory \\
} $^\ast$equal contributors \quad \textsuperscript{\Letter}corresponding author \\
 {\tt\small wychao1987@gmail.com}
\end{center}

\vskip 0.3in
}]

\definecolor{mygreen}{RGB}{93,173,85}
\definecolor{myred}{RGB}{192, 0, 0}
\makeatother
\newcommand{\pub}[1]{\color{gray}{\tiny{[{#1}]}}}
\newcommand{\reshll}[2]{
{#1} \fontsize{7.5pt}{1em}\selectfont\color{mygreen}{$\!\uparrow\!$ {#2}}
}
\newcommand{\reshl}[2]{
\textbf{#1} \fontsize{7.5pt}{1em}\selectfont\color{mygreen}{$\!\uparrow\!$ \textbf{#2}}
}





\begin{abstract}
    Class incremental learning aims to enable models to learn from sequential, non-stationary data streams across different tasks without catastrophic forgetting. 
    In class incremental semantic segmentation (CISS), the semantic content of image pixels evolves over incremental phases, known as \textbf{ semantic drift}. 
    In this work, we identify two critical challenges in CISS that contribute to semantic drift and degrade performance. First, we highlight the issue of separate optimization, where different parts of the model are optimized in distinct incremental stages, leading to misaligned probability scales. Second, we identify noisy semantics arising from inappropriate pseudo-labeling, which results in sub-optimal results.
    To address these challenges, we propose a novel and effective approach, Image Posterior and Semantics Decoupling for Segmentation (IPSeg). IPSeg introduces two key mechanisms: (1) leveraging image posterior probabilities to align optimization across stages and mitigate the effects of separate optimization, and (2) employing semantics decoupling to handle noisy semantics and tailor learning strategies for different semantics. Extensive experiments on the Pascal VOC 2012 and ADE20K datasets demonstrate that IPSeg achieves superior performance compared to state-of-the-art methods, particularly in challenging long-term incremental scenarios. Our code is now available at \href{https://github.com/YanFangCS/IPSeg}{https://github.com/YanFangCS/IPSeg}.
\end{abstract}

\section{Introduction}
\label{sec:Introduction}

\begin{figure*}[ht]
    \centering
    \includegraphics[width=0.9\linewidth]{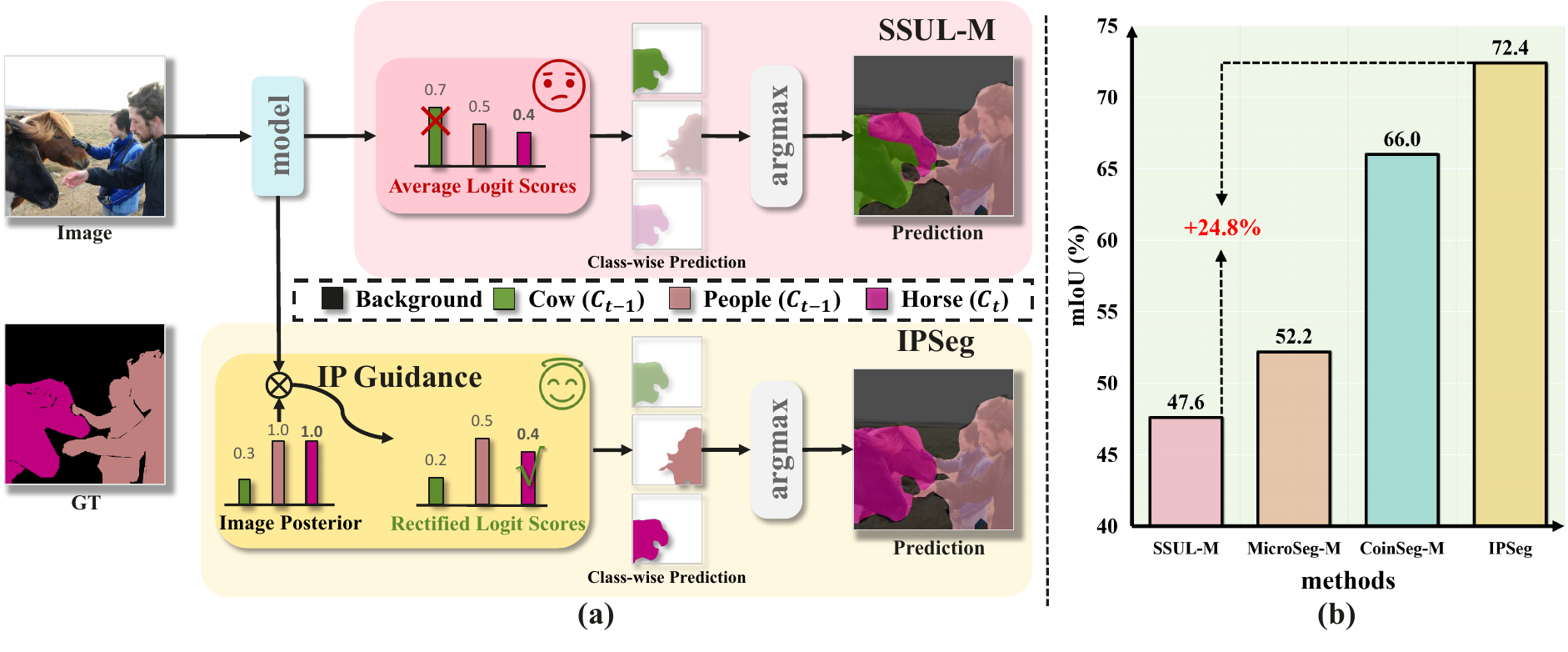}
    \vspace{-10pt}
    \caption{(a) Due to the existence of \textit{separate optimization}, the previous method SSUL-M misclassifies a ``horse'' as a ``cow'' with higher logit scores when learning ``horse'' following ``cow''. 
    While our IPSeg leverages image posterior (IP) guidance to produce accurate predictions on these two similar-look classes.
    The ``logit scores'' refer to pixel-wise prediction, and the image posterior refers to our introduced image-wise prediction.
    The logit numbers are used for better illustration.
    (b) The quantitative performance comparison with state-of-the-art methods under the long-term incremental challenge (VOC 2-2).}
    \label{fig:vis_intro}
\end{figure*}

Deep learning methods have achieved significant success in vision~\citep{Cl_in_cv_qu2021recent} and language~\citep{CL_in_NLP_ke2022continual} tasks with fixed or stationary data distributions. However, real-world scenarios are characterized by dynamic and non-stationary data distributions, posing the challenge of \textit{catastrophic forgetting}~\citep{CF_mccloskey1989catastrophic, mcclelland1995there}. Incremental learning, a.k.a. continual learning or lifelong learning~\citep{lifelong_silver2013lifelong}, has been proposed to enable models to adapt to new data distributions without forgetting previous knowledge~\citep{lifelong2_kudithipudi2022biological}. Within this domain, Class Incremental Learning (CIL) methods~\citep{CIL_archi_serra2018overcoming, lwf_li2017learning, iCaRL_rebuffi2017icarl, OCIL_survey_mai2022online, wang2024hierarchical} have shown great potential in learning new classes from incoming data, particularly for classification tasks~\citep{CL_in_classification_de2021continual}.

Class Incremental Semantic Segmentation (CISS) extends the principles of CIL to pixel-wise tasks.
In addition to catastrophic forgetting, CISS encounters an even more critical challenge: \textit{semantic drift}~\citep{CISS_survey_yuan2023survey} or \textit{background shift}~\citep{MiB_cermelli2020modeling}, which describes the incremental change in the semantic meaning of pixel labels. 
Several studies~\citep{PLOP_douillard2021plop, SSUL_cha2021ssul, microseg_zhang2022mining, coinseg_zhang2023coinseg} attribute \textit{semantic drift}
to the evolving semantic content of the background across incremental stages.
Subsequent works~\citep{MiB_cermelli2020modeling, PLOP_douillard2021plop} early pioneer this investigation using knowledge distillation and pseudo-labeling. 
More recent works~\citep{SSUL_cha2021ssul, microseg_zhang2022mining, coinseg_zhang2023coinseg} further use saliency maps and segment proposals to differentiate between the foreground and background pixels. 
These works introduce naive and inappropriate pseudo-labeling, ignoring decoupling the learning of these semantics, leaving \textit{noisy semantics} remains a critical challenge to be solved.

In this paper, we delve into \textit{semantic drift} challenge and identify an additional but more essential issue, \textit{separate optimization}.
\textit{Separate optimization} refers to the learning manner within CISS methods that independently and sequentially update the task heads for each target class set and freeze previous task heads to prevent catastrophic forgetting.
This leads to a scenario in which different task heads trained in different stages always have misaligned probability scales, especially on similar-looking classes. It finally results in error classification and magnifying semantic drift.
~\cref{fig:vis_intro}(a) directly presents the impact of \textit{separate optimization}, where the SSUL-M model mistakenly classifies a horse as a ``cow'' class with a higher logit score when separately training the corresponding task heads.
Under the combined impacts of \textit{separate optimization} and \textit{noisy semantics}, the previous efforts are still short of effectively addressing the \textit{semantic drift} challenge.

Motivated by our observations and analyses, we introduce \textbf{I}mage \textbf{P}osterior and Semantics Decoupling for Class-Incremental Semantic \textbf{Seg}mentation (IPSeg) to address the aforementioned challenges.
We propose \textbf{image posterior guidance} to mitigate \textit{separate optimization} by rectifying the misaligned pixel-wise predictions using image-wise predictions.
As illustrated in ~\cref{fig:vis_intro}(a), IPSeg correctly predicts ``horse'' with the assistance of image posterior guidance.
Furthermore, we propose \textbf{permanent-temporary semantics decoupling} to decouple \textit{noisy semantics} into two groups, one characterized by simple and stable semantics, and the other by complex and dynamic semantics. We also introduce separate learning strategies for better decoupling.


Extensive experimental results on two popular benchmarks, Pascal VOC 2012 and ADE20K, demonstrate the effectiveness and robustness of IPSeg. Our method consistently outperforms the state-of-the-art methods across various incremental scenarios, particularly in long-term challenges as gaining \(\textbf{24.8}\)\% improvement in VOC 2-2 task. 

\section{Related Work}
\label{sec:Related Work}

\paragraph{Class Incremental Learning (CIL)}
Class-incremental learning is a method that continuously acquires knowledge in the order of classes, aiming to address catastrophic forgetting~\citep{CF_mccloskey1989catastrophic}. Existing work~\citep{CIL_survey_wang2024comprehensive} broadly categorizes these approaches into three main types.
Replay-based methods involve storing data or features of old classes or generating data that includes old classes. They can be further divided into Experience Replay~\citep{iCaRL_rebuffi2017icarl, RM_bang2021rainbow}, Generative Replay~\citep{CIL_G_replay_liu2020generative, CIL_G_replay_shin2017continual}, and Feature Replay~\citep{CIL_F_replay_belouadah2019il2m}.
Regularization-based methods focus on designing loss functions that incorporate second-order penalties based on the contribution of parameters to different tasks~\citep{ewc_kirkpatrick2017overcoming, CIL_loss_jung2020continual}. They also rely on knowledge distillation, using the model from the previous phase as a teacher to constrain current model~\citep{lwf_li2017learning, iCaRL_rebuffi2017icarl, PODnet_douillard2020podnet, DER_buzzega2020dark}.
Architecture-based methods dynamically adjust model parameters based on new data, including assigning specific parameters for different data~\citep{CIL_archi_gurbuz2022nispa, CIL_archi_serra2018overcoming} and breaking down model parameters into task-specific or shared parts~\citep{DyTox_douillard2022dytox}.

\paragraph{Class Incremental Semantic Segmentation (CISS)}
 CISS is similar to class incremental learning (CIL) but extends the task to pixel-level predictions~\citep{reminder_phan2022class,pcss_camuffo2023continual,rcil_zhang2022representation,ewf_xiao2023endpoints}. MiB~\citep{MiB_cermelli2020modeling} first introduces the concept of semantic shift unique to CISS, employing distillation strategies to mitigate this issue. PLOP~\citep{PLOP_douillard2021plop} utilizes pseudo-labeling techniques for incremental segmentation to address background shift, while SSUL~\citep{SSUL_cha2021ssul} further incorporates salient information, introducing the concept of ``unknown classes'' into each learning phase and using a memory pool to store old data to prevent catastrophic forgetting. RECALL~\citep{recall_maracani2021recall} and DiffusePast~\citep{diffusepast_chen2023diffusepast} extend traditional replay methods by incorporating synthetic samples of previous classes generated using Diffusion~\citep{diffusion_ho2020denoising} or GAN~\citep{GAN_goodfellow2020generative} models. MicroSeg~\citep{microseg_zhang2022mining} employs a proposal generator to simulate unseen classes. CoinSeg~\citep{coinseg_zhang2023coinseg} highlights differences within and between classes, designing a contrastive loss to adjust the feature distribution of classes. PFCSS~\citep{PFCSS_lin2023preparing} emphasizes the preemptive learning of future knowledge to enhance the model's discrimination ability between new and old classes.

\section{Method}
\label{sec:Method}


%

In this section, we begin by presenting the necessary notation and definition of the problem, followed by our analysis of \textit{semantic drift} in Section~\ref{sec3-1:preliminary}. Next, we introduce our proposed method, IPSeg, with detailed designs including image posterior and semantics decoupling in Section~\ref{sec3-3} and Section~\ref{sec3-4}.


\subsection{Preliminary}
\label{sec3-1:preliminary}


\paragraph{Notation and problem formulation} Following previous works~\citep{SSUL_cha2021ssul,microseg_zhang2022mining,coinseg_zhang2023coinseg}, 
in CISS, a model needs to learn the target classes $\mathcal{C}_{1:T}$ from a series of incremental tasks as $t=1,2,3,...,T$. For task $t$, the model learns from a unique training dataset $\mathcal{D}_t$ which consists of training data and ground truth pairs $\mathcal{D}_t = \{(x_{i}^t, y_{i}^t)\}_{i=1}^{\left|\mathcal{D}_t\right|}$. Here $i$ denotes the sample index, $t$ for the task index, and $\left|\mathcal{D}_t\right|$ for the training dataset scale. 
$x_{i,j}^t$ and $y_{i, j}^t$ denote the $j$-th pixels and the annotation in the image $x_i^t$.
In each incremental phase $t$, the model can only access the class set $\mathcal{C}_t \cup c_b$ where $\mathcal{C}_t$ denotes the class set of current task $t$ and $c_b$ for background class. 

To prevent catastrophic forgetting, architecture-based methods allocate and optimize distinct sets of parameters for each class, instead of directly updating the whole model $f_t$. Typically, $f_t$ is composed of a frozen backbone $h_\theta$ and a series of learnable task heads $\phi_{1:t}$, with one task head corresponding to a specific task. 
In task $t$, only the new task head $\phi_t$ is set to be optimized.
In inference, the prediction for the $j$-th pixel in image $x_i$ can be obtained by:
\begin{equation}
    \vspace{-10pt}
    \hat{y}_{i,j} = f_{t}(x_{i,j}) = \mathop{\arg\max}\limits_{c\in\mathcal{C}_{1:T}} \phi_{1:T}^c(h_\theta(x_{i,j})).
\end{equation}

Where \(\phi_{1:T}^c(\cdot)\) denotes the $C$-dimension outputs. 
Additionally, we introduce the image-level labels $\mathcal{Y}_i$ of the image $x_i$, a memory buffer $\mathcal{M}$, and an extra image classification head $\psi$ in our implementation. A comprehensive list and explanation of symbols can be found in the appendix.

\paragraph{Semantic Drift}  
Previous work~\citep{ewc_kirkpatrick2017overcoming} mainly attributes the \textit{semantic drift} to \textit{noisy semantics} within the background class $c_b$. They attempt to mitigate this challenge by decoupling the class \( c_b \) into subclasses $c'_b$ and $c_u$, where $c'_b$ denotes the pure background and $c_u$ denotes the unknown class. The most advanced methods~\citep{microseg_zhang2022mining,SSUL_cha2021ssul} further decouple the unknown classes \( c_u \) into past seen classes \( \mathcal{C}_{1:t-1} \) and dummy unknown class \( c'_u \) using pseudo labeling. 
However, \textit{semantic drift} remains unresolved as the decoupled classes are still evolving across incremental phases while the coupled training strategy is not able to cope with noisy pseudo labels.

Additionally, another essential challenge, \textit{separate optimization} inherent within incremental learning also contributes to \textit{semantic drift} but attracts little attention. Recent work~\citep{eclipse_kim2024eclipse} finds a similar phenomenon that freezing parameters from the old stage can preserve the model's prior knowledge but introduces error propagation and confusion between similar classes. 
In architecture-based methods, the task head \( \phi_t \) is exclusively trained by supervision from the current classes and will be frozen to resist catastrophic forgetting in the following incremental phases. In the following task $t_1, t_1 > t$, \( \phi_t \) may predict high scores on objects from other appearance-similar classes, without any penalty and optimization. 
In this incremental learning manner, task heads trained in different stages always have misaligned probability scales, and generate error predictions, especially on similar classes.
This \textit{separate optimization} manner ultimately causes the incremental models to misclassify some categories and makes \textit{semantic drift} more difficult to thoroughly address. In the appendix, some cases can be found to help understand this challenge.


\begin{figure*}[t]
    \centering
    \includegraphics[width=0.9\textwidth]{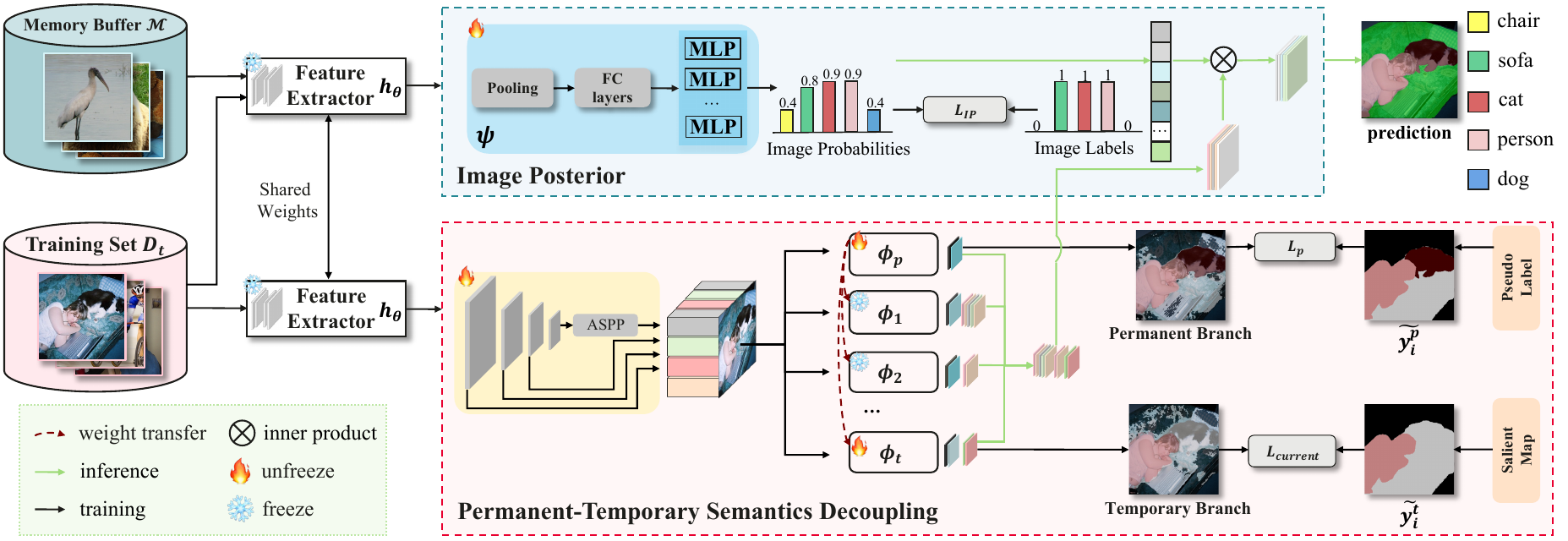}
    \caption{Overall architecture of our proposed IPSeg, mainly composed of image posterior and permanent-temporary semantics decoupling two parts. In the latter part, $\phi_p$ denotes the permanent learning branch and $\phi_1, \phi_2, ..., \phi_t$ for temporary ones. The black solid lines are used to indicate the data flow in training and the green ones are for inference.}
    \label{fig:overview}
\end{figure*}

\subsection{Overview}
\label{sec3-2:overview}

As illustrated in Figure~\ref{fig:overview}, we propose \textbf{I}mage \textbf{P}osterior and Semantics Decoupling for Class-incremental Semantic \textbf{Seg}mentation (IPSeg) to mitigate \textit{semantic drift} through two main strategies: image posterior guidance and permanent-temporary semantics decoupling. In Section~\ref{sec3-3}, we describe how the IPSeg model uses image posterior guidance to mitigate \textit{separate optimization}. 
To address \textit{noisy semantics}, IPSeg employs branches with different learning cycles to decouple the learning of noisy semantics. Detailed explanations of this approach are provided in Section~\ref{sec3-4}.


\subsection{Image Posterior Guidance}
\label{sec3-3}

As previously discussed, the \textit{separate optimization} leads to misaligned probability scales across different incremental task heads and error predictions. 
We propose leveraging the image-level posterior as the global guidance to correct the probability distributions of different task heads. The rationale for using the image posterior probabilities is based on the following fact:

\textbf{Fact}: \textit{For any image, if its image-level class domain is \(\mathcal{C}_I\) and its pixel-level class domain is \(\mathcal{C}_P\), the class domains \(\mathcal{C}_I\) and \(\mathcal{C}_P\) are the same, i.e., \(\mathcal{C}_I = \mathcal{C}_P\).}

Inspired by this fact, we propose to use an extra image posterior branch $\psi$ to predict image classification labels and train it in an incremental learning manner. As illustrated in Figure~\ref{fig:overview}, $\psi$ is composed of Pooling, Fully connected (FC) layers, and Multi-Layer Perceptrons (with one MLP per step) with the input dimension of 4096 and the output dimension of \(\left|\mathcal{C}_{1:T}\right|\), where the FC layers serve as shared intermediate feature processors, and the MLPs serve as incremental classification heads for incremental classes. 



In task \( t \) (\( t > 1 \)), the model can only access data \( x^{m}_i \) from the memory buffer \(\mathcal{M}\) and \( x^{t}_i \) from the current training dataset \(\mathcal{D}_t\). Previous works~\cite{SSUL_cha2021ssul, coinseg_zhang2023coinseg} put \( x^{m}_i \) into the training phase to revisit and reinforce prior knowledge of segmentation by simply rehearsal. IPSeg further takes advantage of the rich class distribution knowledge in \( x^{m}_i \) to train and enhance the image posterior branch.

In IPSeg, the mixed data samples \( x^{m,t}_i \) from  \(\mathcal{M}\) and \(\mathcal{D}_t\) are processed by the network backbone \( h_{\theta} \) into the image feature \( h_{\theta}(x^{m,t}_i) \), and further processed by image posterior branch \(\psi\) into the image classification prediction \(\hat{\mathcal{Y}}_i^{m,t}\). The objective function for training $\psi$ is: 
\vspace{-5pt}
\begin{equation}
    \begin{aligned}
        \mathcal{L}_{\text{\tiny IP}} &= \mathcal{L}_{\text{\tiny BCE}}(\hat{\mathcal{Y}}^{m,t}_i, \tilde{\mathcal{Y}}^{m,t}_i) = \mathcal{L}_{\text{\tiny BCE}}(\psi(h_\theta(x^{m,t}_i)), \tilde{\mathcal{Y}}^{m,t}_i), \\
        \tilde{\mathcal{Y}}^{m,t}_i &= \mathcal{Y}^{m,t}_i \cup \tilde{\mathcal{Y}}_{\phi_{1:t-1}(h_\theta(x^{m,t}_i))}.
    \end{aligned}
\end{equation}
Where image classification label \(\tilde{\mathcal{Y}}^{m,t}_i\) consists of two parts, the ground truth label \(\mathcal{Y}^{m,t}_i\) of the data \( x^{m,t}_i \) and pseudo label \(\tilde{\mathcal{Y}}_{\phi_{1:t-1}(h_\theta(x^{m,t}_i))}\) on past seen classes $\mathcal{C}_{1:t-1}$. 
Instead of relying solely on the label \(\mathcal{Y}^{m,t}_i\), we use the image-level pseudo labels from previous task heads prediction to enhance the model's discriminative ability on prior classes.


During inference, the image posterior branch predicts posterior probabilities on all classes \(\mathcal{C}_{1:T}\). For a testing image \( x_i \), the final pixel-wise scores are computed by element-wise multiplication between the image posterior probabilities from \(\psi\) and the pixel-wise probabilities from \(\phi_{0:T}\): 
\begin{equation}
    p_{i} = 
    \underbrace{\texttt{Concat}(\alpha_{\text{\tiny BC}},\sigma(~\psi(h_\theta(x_i)))) }_{\text{Image Posterior Probability}}
    \cdot  
    \sigma (\underbrace{\phi_{0:T}(h_{\theta}(x_i))}_{\text{Pixel-wise Probability}}).
    \label{equ_3}
\end{equation}

Where \(\sigma(\cdot)\) denotes the Sigmoid function.
The hyperparameter \(\alpha_{\text{\tiny BC}}\) is used to compensate for the lack of background posterior probability, with the default value \(\alpha_{\text{\tiny BC}}=0.9\). 
The result \( p_i \) is the rectified pixel-wise prediction with a shape of \([C, HW]\), and \( p^c_{i,j} \) is prediction of the \( j \)-th pixel on class $c$. The prediction of the $j$-th pixel can be written as:
\begin{equation}
    \hat{y}_{i,j} = \mathop{\arg\max}\limits_{c\in\mathcal{C}_{1:t}} p_{i,j}^c.
    \label{equ_4}
\end{equation}

\subsection{Permanent-Temporary Semantics Decoupling}
\label{sec3-4}

To further address \textit{semantic drift} caused by the coupled learning of complex and noisy pseudo labels \( c_b \) and \( c_u \) along with incomplete yet accurate label \(\mathcal{C}_t\), we propose a decoupling strategy that segregates the learning process for different semantics. Here is our empirical observation:

\textbf{Observation}: \textit{Given an image in incremental task t, the semantic contents of it can be divided into four parts: past classes \(\mathcal{C}_{1:t-1}\), target classes \(\mathcal{C}_{t}\), unknown foreground \(c'_u\) and pure background \(c'_b\).}

Based on this observation, we first introduce dummy label \(c_f=\mathcal{C}_{1:t-1} \cup c'_u\) to represent the foreground regions that encompass both past seen classes and unknown classes, which are not the primary targets in the current task. Subsequently, we decouple the regions of a training image into two sets: \( \mathcal{C}_t  \cup  c_f\) and \( c'_b \cup c'_u \). The former set  \( \mathcal{C}_t  \cup  c_f\) are current target classes and other foreground objects, which are temporary concepts belonging to specific incremental steps, and change drastically as the incremental steps progress. In contrast, \( c'_b \cup c'_u \) are pseudo labels representing pure background and unknown objects, which are permanent concepts, exist across the whole incremental steps and maintain stable (\( c'_b \) remains fixed, \( c'_u \) shrinks but does not disappear). 

The learning of these two sets is also decoupled. The current task head \(\phi_t\) serves as the temporary branch to learn the semantics \( \mathcal{C}_t \) $\cup$ \( c_f \) existing in the current incremental phase. Besides, we introduce a permanent branch \(\phi_p\) to learn the permanent dummy semantics \( c'_b \) and \( c'_u \). \(\phi_p\) has the same network architecture as \(\phi_{t}\). They are composed of three 3x3 convolution layers and several upsampling layers. It's worth noting that $\phi_p$ and $\phi_t$  have different learning cycles as illustrated in Figure~\ref{fig:overview}. The permanent branch $\phi_p$ is trained and optimized across all incremental phases to distinguish unknown objects and the background. While temporary branch $\phi_t$ ($t=1,2,...,T$) is temporarily trained in the corresponding task phase $t$ to recognize target classes $\mathcal{C}_t$.
Following our decoupling strategy, we can reassign the labels of image \( x_i \) as:
\vspace{-10pt}
\begin{equation}
    \scriptsize
    \begin{aligned}
        \tilde{y}^p_{i} &= \begin{cases}
            c_i, & \text{if} ~y^t_{i}\in\mathcal{C}_t \vee \left( (y^t_{i}=c_b) \wedge \left( f_{t-1}(x_i)\in\mathcal{C}_{1:t-1} \right) \right) \\
            c'_u, & \text{if} ~\left( y^t_{i}=c_b \right) \wedge \left( f_{t-1}(x_i) \notin \mathcal{C}_{1:t-1} \right) \wedge \left( S(x_i)=1 \right) \\
            c'_b, & \text{else,}
        \end{cases}, \\
        \tilde{y}^t_{i} &= \begin{cases}
            y^t_{i}, & \text{if} ~y^t_{i}\in\mathcal{C}_t \\
            c_f, & \text{if} ~\left( y^t_{i}=c_b \right) \wedge \left( S(x_i)=1 \right) \\
            c'_b, & \text{else,}
        \end{cases}.
    \end{aligned}
    \normalsize
    \label{equ_label}
\end{equation}
\vspace{-10pt}

Where \(f_{t-1}(\cdot)\) is the model of task $t-1$ and \( S(\cdot) \) is the salient object detector as used in SSUL~\citep{SSUL_cha2021ssul}. \(\tilde{y}_{i}^{p}\) is the label used to train \(\phi_p\), and \(\tilde{y}_{i}^{t}\) is the label used to train \(\phi_t\) for the current task \( t \). \(c_i\) is the ignored region not included in the loss calculation. The visualization of semantics decoupling is provided in the appendix.

The objective functions for these two branches is defined as:
\begin{equation}
    \begin{aligned}
    \mathcal{L}_{p} = \mathcal{L}_{\text{\tiny BCE}}(~\phi_p(h_\theta(x_i^t)), \tilde{y}_i^p ~), \\
    \mathcal{L}_{\text{\tiny current}} = \mathcal{L}_{\text{\tiny BCE}}(~\phi_t(h_\theta(x_i^t)), \tilde{y}_i^t~).
    \end{aligned}
\end{equation}

\vspace{-10pt}
Finally, the total optimization objective function is:
\begin{equation}
    \mathcal{L}_{total}=\mathcal{L}_{\text{\tiny IP}}+\lambda_{1}\mathcal{L}_{\text{\tiny current}}+\lambda_{2}\mathcal{L}_{p},
\end{equation}
where \(\lambda_1\) and \(\lambda_2\) are trade-off hyperparameters to balance different training objective functions.

During inference, as illustrated by the green lines in Figure 2, the permanent branch $\phi_p$ predicts on the background $c_b'$ and unknown objects $c_u'$, with only $c_b'$ used for inference. Meanwhile, the temporary branch $\phi_t$ ($t=1,2,...,T$) predicts for the target classes $\mathcal{C}_t$, the foreground region \( c_f \) and the background $c_b'$, where $\mathcal{C}_t$ and \( c_f \) are used for inference. The pixel-level prediction $\phi_{0:T}(h_\theta(x_i))$ is formulated as:
\begin{equation}
\phi_{0:T}(h_{\theta}(x_i)) = \texttt{Concat}( ~\phi_{p}(h_\theta(x_i))~,~\phi_{1:T}(h_\theta(x_i))).
\end{equation} 
Where $\phi_{p}(h_\theta(x_i))$ and $\phi_{1:T}(h_\theta(x_i))$ represent background prediction from permanent branch and the aggregated foreground predictions from all temporary branches. The pixel-level prediction is then producted by image posterior probability to form the final prediction maps as Eq~\ref{equ_3} and Eq~\ref{equ_4}.

Furthermore, to mitigate the issue of inaccurate predictions on other foreground classes \(c_f\) within each task head $\phi_t$ during inference, we introduce a Noise Filtering trick, filtering out prediction errors associated with \(c_f\). The prediction for the \( j \)-th pixel \(\hat{y}_{i,j}\) is processed as:
\begin{equation}
\hat{y}_{i,j}=\begin{cases}\alpha_{\text{\tiny NF}} \cdot \hat{y}_{i,j}&\text{if} ~max(~p^f_{i,j},~p^c_{i,j}~)=p^f_{i,j}\\\hat{y}_{i,j}&\text{if}~max(~p^f_{i,j},~p^c_{i,j}~)=p^c_{i,j}\end{cases}
\end{equation}
Where \(\alpha_{\text{\tiny NF}}\) is noise filtering term with the default value \(\alpha_{\text{\tiny NF}}=0.4\). And \(p^f_{i,j}\) and \(p^c_{i,j}\) are the \( j \)-th pixel logit outputs on the foreground \(c_f\) and target class $\mathcal{C}_t$ respectively.

\subsection{Improving Memory Buffer}


The memory buffer \(\mathcal{M}\) plays a crucial role in our implementation and we implement the memory buffer based on unbiased learning and storage efficiency. 
IPSeg employs a class-balanced sampling strategy, ensuring the image posterior branch can adequately access samples from all classes. Specifically, given the memory size \(\left|\mathcal{M}\right|\) and the number of already seen classes \(\left|\mathcal{C}_{1:t}\right|\), the sampling strategy ensures there are at least \(\left|\mathcal{M}\right|//\left|\mathcal{C}_{1:t}\right|\) samples for each class.
IPSeg also optimizes the storage cost of $\mathcal{M}$ by only storing image-level labels and object salient masks for samples. Image-level labels are required for the image posterior branch for unbiased classification. While the salient masks split images into background and foreground objects, labeled with 0 and 1 respectively. This simplification mechanism requires less storage cost compared to previous methods that store the whole pixel-wise annotations on all classes. More details can be found in the appendix.

\begin{table*}[t]
    \centering
    \caption{Comparison with state-of-the-art methods on Pascal VOC 2012 dataset across 4 typical incremental scenarios. ``-'' denotes the results are not provided in the original paper. \textsuperscript{†} denotes the result is reproduced using the official code with Swin-B backbone. ``IPSeg w/o M'' denotes the data-free version of IPSeg, which is trained without memory buffer.}
    \resizebox{0.97\linewidth}{!}{
    \begin{tabular}{c|l|c||ccc|ccc|ccc|ccc}
    \toprule
    \multicolumn{2}{c |}{\multirow{2}{*}{Method}}  & \multirow{2}{*}{Backbone} & \multicolumn{3}{c|}{\textbf{VOC 15-5 (2 steps)}} &\multicolumn{3}{c|}{\textbf{VOC 15-1 (6 steps)}}  &\multicolumn{3}{c|}{\textbf{VOC 10-1 (11 steps)}} &\multicolumn{3}{c}{\textbf{VOC 2-2 (10 steps)}} \\
     \multicolumn{2}{c |}{} & & 0-15 & 16-20 & all & 0-15 & 16-20 & all & 0-10 & 11-20 & all & 0-2 & 3-20 & all \\
    \midrule
    \multirow{9}{*}{\rotatebox{90}{Data-free}}
    
    & MiB~\citep{MiB_cermelli2020modeling} & Resnet-101 & 71.8 & 43.3 & 64.7 & 46.2 & 12.9 & 37.9 & 12.3 & 13.1 & 12.7 & 41.1 & 23.4 & 25.9\\
    & SSUL~\citep{SSUL_cha2021ssul} & Resnet-101 & 77.8 & 50.1 & 71.2 & 77.3 & 36.6 & 67.6 & 71.3 & 46.0 & 59.3 & 62.4 & 42.5 & 45.3 \\
    & IDEC~\citep{idec_zhao2023inherit} & Resnet-101 & 78.0 & 51.8 & 71.8 & 77.0 & 36.5 & 67.3 & 70.7 & 46.3 & 59.1 & - & - & -\\

    & PLOP+NeST~\citep{nest_xie2024early}& Resnet-101 & 77.6 & \textbf{55.8} & 72.4 & 72.2 & 33.7 & 63.1 & 54.2 & 17.8 & 36.9 & - & - & - \\
    & LAG~\citep{lag_yuan2024learning} & Resnet-101 & 77.3 & 51.8 & 71.2 & 75.0 & 37.5 & 66.1 & 69.6 & 42.6 & 56.7 & - & - & -\\
    & IPSeg w/o M (ours) & Resnet-101 & \textbf{78.5} & 55.2 & \textbf{72.9} & \textbf{77.4} & \textbf{41.9} & \textbf{68.9} & \textbf{74.9} & \textbf{52.9} & \textbf{64.4} & \textbf{64.7} & \textbf{51.5} & \textbf{53.4} \\
    \cmidrule{2-15}
    & SSUL~\citep{SSUL_cha2021ssul} & Swin-B & 79.7 & 55.3 & 73.9 & 78.1 & 33.4 & 67.5 & 74.3 & 51.0 & 63.2 & 60.3 & 40.6 & 44.0 \\
    & MicroSeg~\citep{microseg_zhang2022mining} & Swin-B & \textbf{81.9} & 54.0 & 75.2 & 80.5 & 40.8 & 71.0 & 73.5 & 53.0 & 63.8 & 64.8 & 43.4 & 46.5 \\
    & PLOP+NeST~\citep{nest_xie2024early} & Swin-B & 80.5 & \textbf{70.8} & \textbf{78.2} & 76.8 & \textbf{57.2} & 72.2 & 64.3 & 28.3 & 47.3 & - & - & -\\
    & IPSeg w/o M (ours) & Swin-B & 81.4 & 62.4 & 76.9 & \textbf{82.4} & 52.9 & \textbf{75.4} & \textbf{80.0} & \textbf{61.2} & \textbf{71.0} & \textbf{72.1} & \textbf{64.5} & \textbf{65.5} \\
    \midrule
    \multirow{12}{*}{\rotatebox{90}{Replay}} 
    & \textcolor{gray}{Joint} & \textcolor{gray}{Resnet-101} & \textcolor{gray}{80.5} & \textcolor{gray}{73.0} & \textcolor{gray}{78.2} & \textcolor{gray}{80.5} & \textcolor{gray}{73.0} & \textcolor{gray}{78.2} & \textcolor{gray}{79.1} & \textcolor{gray}{77.1} & \textcolor{gray}{78.2} & \textcolor{gray}{73.9} & \textcolor{gray}{78.9} & \textcolor{gray}{78.2}\\
    & SDR~\citep{sdr_michieli2021continual} & Resnet-101 & 75.4 & 52.6 & 69.9 & 44.7 & 21.8 & 39.2 & 32.4 & 17.1 & 25.1 & 13.0 & 5.1 & 6.2\\
    & PLOP-M~\citep{PLOP_douillard2021plop} & Resnet-101 & 78.5 & 65.6 & 75.4 & 71.1 & 52.6 & 66.7 & 57.9 & 51.6 & 54.9 & - & - & -\\
    & SSUL-M~\citep{SSUL_cha2021ssul} & Resnet-101 & 79.5 & 52.9 & 73.2 & 78.9 & 43.9 & 70.6 & 74.8 & 48.9 & 65.5 & 58.8 & 45.8 & 47.6\\
    & MicroSeg-M~\citep{microseg_zhang2022mining} & Resnet-101 & \textbf{82.0} & 59.2 & 76.6 & \textbf{81.3} & 52.5 & 74.4 & \textbf{77.2} & 57.2 & 67.7 & 60.0 & 50.9 & 52.2\\
    & PFCSS-M~\citep{PFCSS_lin2023preparing} & Resnet-101 & 79.9 & 70.2 & 77.1 & 77.1 & \textbf{60.4} & 73.1 & 69.5 & 63.2 & 66.5 & - & - & -\\
    & Adapter~\citep{adapter_zhu2024adaptive} & Resnet-101 & - & - & - & 79.9 & 51.9 & 73.2 & 74.9 & 54.3 & 65.1 & 62.8 & 57.9 & 58.6 \\
    & IPSeg (ours) & Resnet-101 & 79.5 & \textbf{71.0} & \textbf{77.5} & 79.6 & 58.9 & \textbf{74.7} & 75.9 & \textbf{66.4} & \textbf{71.4} & \textbf{62.4} & \textbf{61.0} & \textbf{61.2}\\
    \cmidrule{2-15}
    & \textcolor{gray}{Joint} & \textcolor{gray}{Swin-B} & \textcolor{gray}{83.8} & \textcolor{gray}{79.3} & \textcolor{gray}{82.7} & \textcolor{gray}{83.8} & \textcolor{gray}{79.3} & \textcolor{gray}{82.7} & \textcolor{gray}{82.4} & \textcolor{gray}{83.0} & \textcolor{gray}{82.7} & \textcolor{gray}{75.8} & \textcolor{gray}{83.9} & \textcolor{gray}{82.7}\\
    & SSUL-M\textsuperscript{†}~\citep{SSUL_cha2021ssul} & Swin-B & 79.3 & 55.1 & 73.5 & 78.8 & 49.7 & 71.9 & 75.3 & 54.1 & 65.2 & 61.1 & 47.5 & 49.4\\
    & MicroSeg-M\textsuperscript{†}~\citep{microseg_zhang2022mining} & Swin-B & 82.9 & 60.1 & 77.5 & 82.0 & 47.3 & 73.3 & 78.9 & 59.2 & 70.1 & 62.7 & 51.4 & 53.0\\
    & CoinSeg-M~\citep{coinseg_zhang2023coinseg} & Swin-B & \textbf{84.1} & 69.9 & 80.8 & \textbf{84.1} & 65.5 & 79.6 & \textbf{81.3} & 64.4 & 73.7 & 68.4 & 65.6 & 66.0\\
    & IPSeg (ours) & Swin-B & 83.3 & \textbf{73.3} & \textbf{80.9} & 83.5 & \textbf{75.1} & \textbf{81.5} & 80.3 & \textbf{76.7} & \textbf{78.6} & \textbf{73.1} & \textbf{72.3} & \textbf{72.4}\\
    \bottomrule
    \end{tabular}
    }
    \label{tab:voc_res}
\end{table*}

\section{Experiments}
\label{sec:Experiments}

\subsection{Experimental Setups}
\label{sec4-1:exp_setup}

\paragraph{Dataset} 
Following previous works~\citep{coinseg_zhang2023coinseg,SSUL_cha2021ssul}, we evaluate our method using the Pascal VOC 2012~\citep{voc} and ADE20K~\citep{ade20k} datasets. Pascal VOC 2012 includes 20 foreground classes and one background class, with 10,582 training images and 1,449 validation images. ADE20K, a larger-scale dataset, comprises 150 classes of stuff and objects, with 20,210 training images and 2,000 validation images.

\paragraph{Protocols} 
We primarily use the \textit{overlap} setting to evaluate our method. This setting is more challenging and realistic than the \textit{disjoint} setting~\citep{sats_qiu2023sats}, as the images may contain both seen and unseen classes across different incremental steps. We evaluate IPSeg under several incremental scenarios, denoted as $M$-$N$, where $M$ is the number of classes learned initially, and $N$ is the number of classes learned in each incremental step. For example, VOC 15-1 (6 steps) means learning 15 classes initially and one new class in each subsequent step until all 20 classes are learned. We use the mean Intersection over Union (mIoU) as the evaluation metric.

\paragraph{Implementation details}
Following previous works~\citep{sppa_lin2022continual,DKD_baek2022decomposed,ewf_xiao2023endpoints}, IPSeg utilizes DeepLab V3~\citep{deeplab_v3_chen2017rethinking} as the segmentation model with ResNet-101~\citep{resnet_he2016deep} and Swin Transformer-base (Swin-B)~\citep{swin_liu2021swin} pre-trained on ImageNet-1K~\citep{imagenet_1k_deng2009imagenet} as the backbones. The training batch size is 16 for Pascal VOC 2012 and 8 for ADE20K. IPSeg uses the SGD optimizer with a momentum of 0.9 and a weight decay of $1e$-$4$. The learning rates for both datasets are set to 0.01, with learning rate policies of poly for Pascal VOC 2012 and warm poly for ADE20K. All experiments are conducted with 2 NVIDIA GeForce RTX 3090 GPUs. For a fair comparison, the memory size is set as the same as SSUL~\citep{SSUL_cha2021ssul} that \(|\mathcal{M}|=100\) for Pascal VOC 2012 and \(|\mathcal{M}|=300\) for ADE20K.
Following SSUL~\citep{SSUL_cha2021ssul}, IPSeg is trained for 50 epochs on Pascal VOC 2012 and 60 epochs on ADE20K.
Pseudo-label~\citep{coinseg_zhang2023coinseg} and saliency information~\citep{SOD_deep_used_hou2017deeply} are adopted as previous methods~\citep{SSUL_cha2021ssul, microseg_zhang2022mining}. To avoid information leaking, the ground truth of the training data in the image posterior branch only consists of annotations and pseudo-labels from the corresponding steps.

\paragraph{Baselines} We compare IPSeg with various CISS methods, including MiB~\citep{MiB_cermelli2020modeling}, SDR~\citep{sdr_michieli2021continual}, and PLOP~\citep{PLOP_douillard2021plop}, as well as state-of-the-art methods such as SSUL~\citep{SSUL_cha2021ssul}, MicroSeg~\citep{microseg_zhang2022mining}, PFCSS~\citep{PFCSS_lin2023preparing}, CoinSeg~\citep{coinseg_zhang2023coinseg}, NeST~\citep{nest_xie2024early}, LAG~\citep{lag_yuan2024learning}, and Adapter~\citep{adapter_zhu2024adaptive}. Among these, PFCSS, CoinSeg, and Adapter are the current state-of-the-art replay methods. For a fair comparison, we reproduce some works using their official code with the Swin-B backbone. Additionally, we provide the results of \textbf{Joint} as a theoretical upper bound for incremental tasks. We report incremental results in three parts: initial classes, new classes, and overall classes.

\subsection{Main Results}

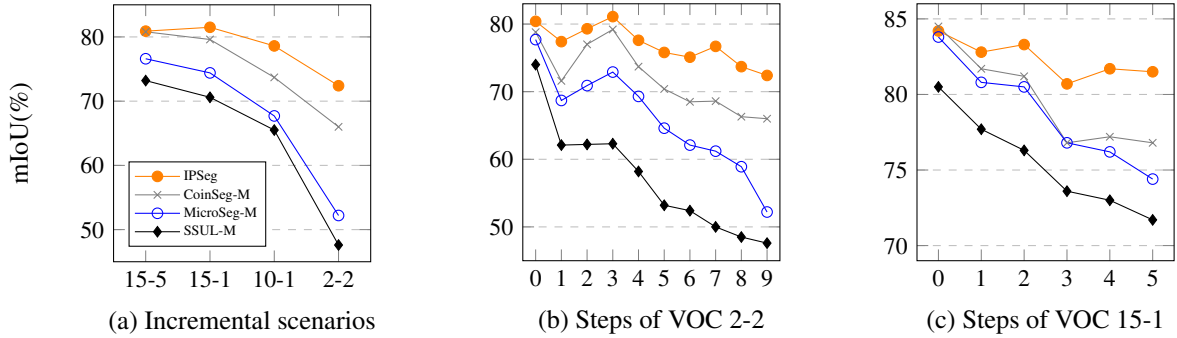
\begin{figure*}[ht]
    \begin{minipage}[t]{0.37\textwidth}
        \centering
        \begin{tikzpicture}
            \begin{axis}[
                xlabel={(a) Incremental scenarios},
                ylabel={mIoU(\%)},
                xmin=0.5, xmax=4.5,
                ymin=45, ymax=85,
                xtick={1,2,3,4},
                xticklabels={\text{\small 15-5},\text{\small 15-1},\text{\small 10-1},\text{\small 2-2}},
                ytick={50, 60, 70, 80},
                yticklabels = {\text{\small 50}, \text{\small 60}, \text{\small 70}, \text{\small 80}},
                legend style={
                    font=\small, 
                    at={(0.05,0.05)}, 
                    anchor=south west,
                    inner sep=2pt, 
                    outer sep=1pt, 
                    nodes={scale=0.6, transform shape} 
                },
                ymajorgrids=true,
                grid style=dashed,
                legend cell align={left},
                width=5cm, 
                height=5cm 
            ]
            
            \addplot[
                color=orange,
                mark=*, 
                mark size=2pt 
                ]
                coordinates {
                (1,80.9)(2,81.5)(3,78.6)(4,72.4)
                };
                \addlegendentry{IPSeg}
            
            \addplot[
                color=gray,
                mark=x,
                mark size=2pt 
                ]
                coordinates {
                (1,80.8)(2,79.6)(3,73.7)(4,66.0)
                };
                \addlegendentry{CoinSeg-M}
            
            \addplot[
                color=blue,
                mark=o,
                mark size=2pt 
                ]
                coordinates {
                (1,76.6)(2,74.4)(3,67.7)(4,52.2)
                };
                \addlegendentry{MicroSeg-M}
            
            \addplot[
                color=black,
                mark=diamond*,
                mark size=2pt 
                ]
                coordinates {
                (1,73.2)(2,70.6)(3,65.5)(4,47.6)
                };
                \addlegendentry{SSUL-M}
            
            
            \end{axis}
        \end{tikzpicture}
    \end{minipage}
    \begin{minipage}[t]{0.3\textwidth}
        \centering
        \begin{tikzpicture}
            \begin{axis}[
                xlabel={(b) Steps of VOC 2-2},
                xmin=-0.5, xmax=9.5,
                ymin=45, ymax=83,
                xtick={0,1,2,3,4,5,6,7,8,9},
                xticklabels={\text{\small 0}, \text{\small 1}, \text{\small 2}, \text{\small 3}, \text{\small 4}, \text{\small 5}, \text{\small 6}, \text{\small 7}, \text{\small 8}, \text{\small 9}},
                ytick={50, 60, 70, 80},
                yticklabels = {\text{\small 50}, \text{\small 60}, \text{\small 70}, \text{\small 80}},
                legend style={
                    font=\small, 
                    at={(0.05,0.05)}, 
                    anchor=south west,
                    inner sep=2pt,
                    outer sep=1pt,
                    nodes={scale=0.6, transform shape}
                },
                ymajorgrids=true,
                grid style=dashed,
                legend cell align={left},
                width=5cm, 
                height=5cm
            ]
            
            \addplot[
                color=orange,
                mark=*, 
                mark size=2pt
                ]
                coordinates {
                (0,80.4)(1,77.4)(2,79.3)(3,81.1)(4,77.6)(5,75.8)(6,75.1)(7,76.7)(8,73.7)(9,72.4)
                };
            
            \addplot[
                color=gray,
                mark=x,
                mark size=2pt
                ]
                coordinates {
                (0,78.8)(1,71.6)(2,77.0)(3,79.2)(4,73.7)(5,70.4)(6,68.5)(7,68.6)(8,66.3)(9,66.0)
                };
            
            \addplot[
                color=blue,
                mark=o,
                mark size=2pt
                ]
                coordinates {
                (0,77.7)(1,68.7)(2,70.9)(3,72.9)(4,69.3)(5,64.6)(6,62.1)(7,61.2)(8,58.9)(9,52.2)
                };
            
            \addplot[
                color=black,
                mark=diamond*,
                mark size=2pt
                ]
                coordinates {
                (0,74.0)(1,62.1)(2,62.2)(3,62.3)(4,58.2)(5,53.2)(6,52.4)(7,50.0)(8,48.5)(9,47.6)
                };
            
            \end{axis}
        \end{tikzpicture}
    \end{minipage}
    \begin{minipage}[t]{0.3\textwidth}
        \centering
        \begin{tikzpicture}
            \begin{axis}[
                xlabel={(c) Steps of VOC 15-1},
                xmin=-0.5, xmax=5.5,
                ymin=69, ymax=86,
                xtick={0,1,2,3,4,5},
                xticklabels={\text{\small 0}, \text{\small 1}, \text{\small 2}, \text{\small 3}, \text{\small 4}, \text{\small 5}},
                ytick={70,75,80,85},
                yticklabels = {\text{\small 70}, \text{\small 75}, \text{\small 80}, \text{\small 85}},
                legend style={
                    font=\small, 
                    at={(0.05,0.05)}, 
                    anchor=south west,
                    inner sep=2pt,
                    outer sep=1pt,
                    nodes={scale=0.6, transform shape}
                },
                ymajorgrids=true,
                grid style=dashed,
                legend cell align={left},
                width=5cm, 
                height=5cm
            ]
            
            \addplot[
                color=orange,
                mark=*, 
                mark size=2pt
                ]
                coordinates {
                (0,84.2)(1,82.8)(2,83.3)(3,80.7)(4,81.7)(5,81.5)
                };
            
            \addplot[
                color=gray,
                mark=x,
                mark size=2pt
                ]
                coordinates {
                (0,84.5)(1,81.7)(2,81.2)(3,76.8)(4,77.2)(5,76.8)
                };
            
            \addplot[
                color=blue,
                mark=o,
                mark size=2pt
                ]
                coordinates {
                
                (0,83.8)(1,80.8)(2,80.5)(3,76.8)(4,76.2)(5,74.4)
                };
            
            \addplot[
                color=black,
                mark=diamond*,
                mark size=2pt
                ]
                coordinates {
                (0,80.5)(1,77.7)(2,76.3)(3,73.6)(4,73.0)(5,71.7)
                };
            \end{axis}
        \end{tikzpicture}
    \end{minipage}
    \caption{(a) The overall performance of different methods on Pascal VOC 2012 under 4 scenarios, (b) mIoU visualization on Pascal VOC 2012 2-2, (c) mIoU visualization on Pascal VOC 2012 15-1.}
    \label{fig:performance_voc}
    \vspace{-5pt}
\end{figure*}

IPSeg is initially designed with a memory buffer \(\mathcal{M}\), enabling it to fully leverage category distribution knowledge from previous samples to mitigate the separate optimization. Consequently, our primary objective is to demonstrate the superiority of IPSeg by comparing it with other replay-based methods, such as ``CoinSeg-M''. Additionally, we also present the results of IPSeg without \(\mathcal{M}\) (denotes as ``IPSeg w/o M'') to show the potential and robustness of IPSeg.

\begin{table*}[t]
    \centering
        \caption{Comparison with state-of-the-art methods on ADE20K dataset. \textsuperscript{†} denotes the result is reproduced using the official code with Swin-B backbone. \textsuperscript{*} denotes the results from a longer training schedule of 100 epochs, while 60 epochs in ours.}

    \resizebox{0.97\linewidth}{!}{
    \begin{tabular}{c|l|c||ccc|ccc|ccc|ccc}
    \toprule
    \multicolumn{2}{c |}{\multirow{2}{*}{Method}}  & \multirow{2}{*}{Backbone} & \multicolumn{3}{c|}{\textbf{ADE 100-5 (11 steps)}} &\multicolumn{3}{c|}{\textbf{ADE 100-10 (6 steps)}}  &\multicolumn{3}{c|}{\textbf{ADE 100-50 (2 steps)}} &\multicolumn{3}{c}{\textbf{ADE 50-50 (3 steps)}} \\
     \multicolumn{2}{c |}{} & & 0-100 & 101-150 & all & 0-100 & 101-150 & all & 0-100 & 101-150 & all & 0-50 & 51-150 & all \\
    \midrule
    \multirow{12}{*}{\rotatebox{90}{Data-free}} 
   
    & MiB~\citep{MiB_cermelli2020modeling} & Resnet-101 & 36.0 & 5.7 & 26.0 & 38.2 & 11.1 & 29.2 & 40.5 & 17.2 & 32.8 & 45.6 & 21.0 & 29.3\\
    & SSUL~\citep{SSUL_cha2021ssul} & Resnet-101&39.9&17.4&32.5&40.2&18.8&33.1&41.3&18.0&33.6&48.4&20.2&29.6 \\
    & MicroSeg~\citep{microseg_zhang2022mining} & Resnet-101 & 40.4&20.5&33.8&\textbf{41.5}&21.6&34.9&40.2&18.8&33.1&48.6&24.8&32.9\\
    & IDEC~\citep{idec_zhao2023inherit} & Resnet-101 & 39.2 & 14.6 & 31.0 & 40.3 & 17.6 & 32.7 & 42.0 & 18.2 & 34.1 & 47.4 & 26.0 & 33.1\\
    & AWT+MiB~\citep{goswami2023attribution} & Resnet-101 & 38.6 & 16.0 & 31.1 & 39.1 & 21.4 & 33.2 & 40.9 & \textbf{24.7} & 35.6 & 46.6 & 27.0 & 33.5 \\
    

    & PLOP+NeST~\citep{nest_xie2024early} & Resnet-101 & 39.3 & 17.4 & 32.0 & 40.9 & 22.0 & 34.7 & \textbf{42.2} & 24.3 & \textbf{36.3} & \textbf{48.7} & \textbf{27.7} & \textbf{34.8} \\
    & BAM~\citep{bam_zhang2025background} & Resnet-101 & 40.5 & 21.1 & 34.1 & 41.1 & 23.1 & 35.2 & 42.0 & 23.0 & 35.7 & 47.9 & 26.5 & 33.7 \\
    & IPSeg w/o M (ours) & Resnet-101 & \textbf{41.0} & \textbf{22.4} & \textbf{34.8} & 41.0 & \textbf{23.6} & \textbf{35.3} & 41.3 & 24.0 & 35.5 & 46.7 & 26.2 & 33.1 \\

    \cmidrule{2-15}
    & SSUL\textsuperscript{†}~\citep{SSUL_cha2021ssul} & Swin-B & 41.3&16.0&32.9&40.7&19.0&33.5&41.9&20.1&34.6&49.5&21.3&30.7\\
    & CoinSeg~\citep{coinseg_zhang2023coinseg} & Swin-B & \textbf{43.1}&24.1&36.8&42.1&24.5&36.2&41.6&26.7&36.6&49.0&28.9&35.6\\
    & PLOP+NeST~\citep{nest_xie2024early} & Swin-B & 39.7 & 18.3 & 32.6 & 41.7 & 24.2 & 35.9 & \textbf{43.5} & 26.5 & 37.9 & \textbf{50.6} & 28.9 & 36.2 \\
    & IPSeg w/o M (ours) & Swin-B & \textbf{43.1} & \textbf{26.2} & \textbf{37.6} & \textbf{42.5} & \textbf{27.8} & \textbf{37.6} & 43.2 & \textbf{29.0} & \textbf{38.4} & 49.3 & \textbf{33.0} & \textbf{38.5} \\

    \midrule
    \multirow{9}{*}{\rotatebox{90}{Replay}} 
     & \textcolor{gray}{Joint} & \textcolor{gray}{Resnet-101} & \textcolor{gray}{43.5} & \textcolor{gray}{29.4} & \textcolor{gray}{38.3} & \textcolor{gray}{43.5} & \textcolor{gray}{29.4} & \textcolor{gray}{38.8} & \textcolor{gray}{43.5} & \textcolor{gray}{29.4} & \textcolor{gray}{38.8} & \textcolor{gray}{50.3} & \textcolor{gray}{32.7} & \textcolor{gray}{38.8}\\
    & SSUL-M~\citep{SSUL_cha2021ssul}& Resnet-101 & \textbf{42.9} & 17.8 & 34.6 & \textbf{42.9} & 17.7 & 34.5 & 42.8 & 17.5 & 34.4 & 49.1 & 20.1 & 29.8 \\
    & TIKP~\citep{TIKP_yu2024tikp} & Resnet-101 & 37.5 & 17.6 & 30.9 & 41.0 & 19.6 & 33.8 & 42.2 & 20.2 & 34.9 & 48.8 & 25.9 & 33.6 \\
    & Adapter\textsuperscript{*}~\citep{adapter_zhu2024adaptive} & Resnet-101 & 42.6 & 18.0 & 34.5 & \textbf{42.9} & 19.9 & 35.3 & \textbf{43.1} & 23.6 & \textbf{36.7} & \textbf{49.3} & \textbf{27.3} & \textbf{34.7}\\
    & IPSeg (ours) & Resnet-101 & 42.4 & \textbf{22.7} & \textbf{35.9} & 42.1 & \textbf{22.3} & \textbf{35.6} & 41.7 & \textbf{25.2} & 36.3 & 47.3 & 26.7 & 33.6 \\
    \cmidrule{2-15}
    & \textcolor{gray}{Joint} & \textcolor{gray}{Swin-B} & \textcolor{gray}{47.2} & \textcolor{gray}{31.9} & \textcolor{gray}{42.1} & \textcolor{gray}{47.2} & \textcolor{gray}{31.9} & \textcolor{gray}{42.1} & \textcolor{gray}{47.2} & \textcolor{gray}{31.9} & \textcolor{gray}{42.1} & \textcolor{gray}{54.6} & \textcolor{gray}{35.5} & \textcolor{gray}{42.1}\\
    & SSUL-M\textsuperscript{†}~\citep{SSUL_cha2021ssul} & Swin-B & 41.6 & 20.1 & 34.5 & 41.6 & 19.9 & 34.4 & 41.5 & 48.0 & 33.7 & 47.6 & 18.8 & 28.5\\
    & CoinSeg-M\textsuperscript{†}~\citep{coinseg_zhang2023coinseg} & Swin-B & 42.8 & 24.8 & 36.8 & 39.6 & 24.8 & 34.7 & 38.7 & 23.7 & 33.7 & 48.8 & 28.9 & 35.4\\
    & IPSeg (ours) & Swin-B & \textbf{43.2} & \textbf{30.4} & \textbf{38.9} & \textbf{43.0} & \textbf{30.9} & \textbf{39.0} & \textbf{43.8} & \textbf{31.5} & \textbf{39.7} & \textbf{51.1} & \textbf{34.8} & \textbf{40.3}\\
    \bottomrule
    \end{tabular}
    }
    \label{tab:ade_res}
\end{table*}

\paragraph{Results on Pascal VOC 2012} 
We evaluate IPSeg in various incremental scenarios on Pascal VOC 2012, including standard incremental scenarios (15-5 and 15-1) and long-term incremental scenarios (10-1 and 2-2). As shown in Table~\ref{tab:voc_res}, among the replay-based methods, IPSeg achieves the best results across all incremental scenarios on Pascal VOC 2012 with both ResNet-101 and Swin-B backbones. Notably, in the long-term incremental scenarios 10-1 and 2-2, IPSeg achieves performance gains of $\textbf{4.9}$\% and $\textbf{6.4}$\% over the second-best method, CoinSeg-M, with the same Swin-B backbone.
Meanwhile, the data-free version of IPSeg (denotes as ``IPSeg w/o M'' ) also demonstrates competitive performance, though without specialized designs.

The superior and robust performance of IPSeg is mainly attributed to the reliable role of guidance provided by the image posterior branch.
The image posterior design effectively helps IPSeg avoid catastrophic forgetting and achieve excellent performance on new classes, which often suffer from semantic drift due to separate optimization in new steps.
Additionally, the semantics decoupling design enables IPSeg to better learn foreground classes within each incremental step. These designs bring the improvement of $\textbf{12.3}$\% and $\textbf{6.7}$\% over CoinSeg-M on new classes (11-20) in the 10-1 and 2-2 scenarios, respectively.


Furthermore, as illustrated in Figure~\ref{fig:performance_voc}(a), IPSeg experiences less performance degradation compared to previous state-of-the-art methods as the number of incremental steps increases, which indicates that IPSeg has stronger resistance to catastrophic forgetting. This conclusion is further supported by the data in Figure~\ref{fig:performance_voc}(b) and Figure~\ref{fig:performance_voc}(c), which show that IPSeg exhibits minimal performance declines as the incremental process continues. In contrast, other methods only maintain comparable performance during the initial incremental learning step but quickly degrade in subsequent steps due to catastrophic forgetting. This detailed trend of performance decline across steps validates the effectiveness and robustness of IPSeg in resisting forgetting.

\begin{table*}[t]
    \caption{Ablation on different label choices to incrementally train the image posterior branch.}
    \centering
    \resizebox{0.97\linewidth}{!}{
    \begin{tabular}{l||c|c||ccc|ccc|ccc}
    \toprule
    \multirow{2}{*}{Methods} & \multicolumn{2}{c||}{Labels} & \multicolumn{3}{c|}{\textbf{VOC 15-5 (2 steps)}} &\multicolumn{3}{c|}{\textbf{VOC 10-1 (11 steps)}} &\multicolumn{3}{c}{\textbf{VOC 2-2 (10 steps)}} \\
    & \(\mathcal{C}_{1:t-1}\) & \(\mathcal{C}_{t}\) & 0-15 & 16-20 & all & 0-10 & 11-20 & all & 0-2 & 3-20 & all \\
    \midrule
    Part-GT & \XSolidBrush & GT & 83.3 & 72.8 & 80.8 & 79.3 & 74.5 & 77.0 & 72.6 & 69.4 & 69.8\\
    Pseudo (ours) & PL & GT & 83.3 & 73.3 & 80.9 & 80.3 & 76.7 & 78.6 & 73.1 & 72.3 & 72.4 \\
    Full-GT & GT & GT & 83.2 & 73.8 & 81.0 & 80.1 & 78.0 & 79.2 & 75.2 & 74.6 & 74.8 \\
    \bottomrule
    \end{tabular}
    }
    \label{tab:impact_image-level_pseudo_label}
\end{table*}

\paragraph{Results on ADE20K}
We also conduct a comparison between IPSeg and its competitors under different incremental scenarios on the more challenging ADE20K dataset with two backbones. As shown in Table~\ref{tab:ade_res}, IPSeg consistently achieves performance advantages compared with other replay-based methods, which is similar to those observed on Pascal VOC 2012. Notably, IPSeg with Swin-B backbone demonstrates more significant improvements over its competitors across all incremental scenarios on the ADE20K dataset, with the smallest improvement of $\textbf{2.1}\%$ in the 100-5 scenario and the largest improvement of $\textbf{6.0}\%$ in the 100-50 scenario. The superior performance on the more realistic and complex ADE20K dataset further demonstrates the effectiveness and robustness of IPSeg. 

    
    

\subsection{Ablation Study}
\label{sec4-3:ablations}

\begin{table}[t]
    \caption{Overall ablation study for IPSeg on VOC 15-1.}
    \centering
    \resizebox{0.93\linewidth}{!}{
    \begin{tabular}{c c c | ccc}
        \toprule
        \multirow{2}{*}{IP} & \multirow{2}{*}{SD} & \multirow{2}{*}{NF} & \multicolumn{3}{c}{\textbf{VOC 15-1 (6 steps)}} \\
        & & & 0-15 & 16-20 & all \\
        
        \midrule
        \XSolidBrush & \XSolidBrush & \XSolidBrush  & 78.8 & 49.7 & 71.9 \\
        \Checkmark & \XSolidBrush & \XSolidBrush & 79.4 & 69.6 &\reshll{77.0}{5.1} \\
        \XSolidBrush & \Checkmark & \XSolidBrush& 83.1 & 65.1 &\reshll{78.8}{6.9} \\
        \Checkmark & \Checkmark & \XSolidBrush  & 83.4 & 74.7 &\reshll{81.3}{9.4}  \\
        
        \Checkmark & \Checkmark & \Checkmark  & \textbf{83.6} & \textbf{75.1} & \reshl{81.6}{9.7}  \\
        \bottomrule
    \end{tabular}
    }
    \label{tab:abs_overall}
    \vspace{-10pt}
\end{table}


\paragraph{Ablation on IP branch} The image posterior (IP) branch is trained incrementally but faces challenges due to the lack of labels for old classes. To address this issue, we employ image-level pseudo-label \(\tilde{\mathcal{Y}}^{m,t}_i\) (PL) instead of directly using the partial ground truth label \(\mathcal{Y}^{m,t}_i\), providing comprehensive supervision at the risk of introducing noise due to the inconsistencies between previous heads predictions and current training labels. As shown in Table~\ref{tab:impact_image-level_pseudo_label}, our method achieves significant improvement compared to using only partial ground truth (Part-GT),
and narrows the gap with the upper bound (Full-GT).
This indicates that using \(\tilde{\mathcal{Y}}^{m,t}_i\) is an efficient trade-off, where the benefits of additional supervision from pseudo labels outweigh the potential noise. With this training design, the image posterior branch helps IPSeg effectively mitigate separate optimization and shows superior performance.

\vspace{-8pt}
\paragraph{Ablation on proposed components} 
We analyze the effect of the components in IPSeg, including Image Posterior (\textbf{IP}), Semantics Decoupling (\textbf{SD}), and the Noise Filtering (\textbf{NF}) in SD. All ablations are implemented in Pascal VOC 2012 using the Swin-B backbone. As shown in Table~\ref{tab:abs_overall}, the second row indicates that \textbf{IP} brings significant improvement to new classes. Benefiting from \textbf{IP}'s ability to align probability scales between different task heads, the reliable guidance prevents model performance from degradation caused by \textit{separate optimization}. The third row shows the excellent ability of \textbf{SD} in learning foreground targets at each step by effective decoupling. 
The fourth and fifth rows demonstrate IPSeg's outstanding performance on both old and new classes using \textbf{IP} and \textbf{SD} together. More ablation studies and visualization results for IPSeg are provided in the appendix. 

\section{Conclusions and Limitations}
\label{sec:Conclusions}
In this paper, We propose IPSeg, a simple yet effective method designed to address the issue of semantic drift in class incremental semantic segmentation. We begin by analyzing the details of semantic drift, identifying two key issues: separate optimization and noisy semantics. To mitigate these issues, IPSeg introduces two specific designs: image posterior guidance and semantics decoupling. Experimental results on the Pascal VOC 2012 and ADE20K datasets demonstrate the superior performance of our method, particularly in long-term incremental scenarios.

\textit{Limitations} While IPSeg introduces a novel and promising approach for the class incremental semantic segmentation challenge, We have to claim that IPSeg is based upon the memory buffer for improving classification ability. The basis limits its potential in privacy-sensitive scenarios. Eliminating the need for a memory buffer and extending IPSeg to a wider range of applications are our future targets.

\section*{Impact Statement} Though IPSeg exhibits superior performance and properties of learning plasticity and memory stability, we realize that the usage of memory buffer leaves a lot of room for discussion. On the one hand, the use of memory buffers brings additional storage costs and the risk of information leakage. On the other hand, the use of memory buffers is related to privacy issues in some cases, such as storing private information without approval. These issues need to be treated with caution in artificial intelligence applications.

\nocite{langley00}
\bibliography{example_paper}
\bibliographystyle{icml2025}

\newpage
\appendix
\onecolumn
\newpage
\section{Appendix}
\subsection{Symbols and Explanations}
Table~\ref{tab:symbol} provides key symbols used in our paper along with their explanations to facilitate a better understanding.
\begin{table}[h]
    \caption{Symbols and explanations}
    \centering
    \renewcommand{\arraystretch}{1.3}
    \begin{tabular}{c c | l}
        \toprule
        \multicolumn{2}{c|}{Symbol} & Explanations\\
        \midrule
        \multirow{6}{*}{\rotatebox{90}{Model architecture}} 
        &\(f_t\) & Model of task $t$. \\
        &\(h_{\theta}\) & The backbone extracting features. \\
        &\(\phi_{1:t}\) & All segmentation heads, outputting the final results. \\
        &\(\phi_t\) & The head of task $t$, representing the temporary branch. \\
        &\(\phi_p\) & The permanent branch. \\
        &\(\psi\) & The image posterior branch. \\
        \midrule
        \multirow{8}{*}{\rotatebox{90}{Semantic concepts}} 
        &\(\mathcal{C}_t\) & The target classes set of task $t$.\\
        &\(\mathcal{C}_{1:t-1}\) & The old classes set. \\
        &\(c'_b\) & The pure background. \\
        &\(c'_u\) & Unknown foreground. \\
        & \(c_i\) & The ignored region that does not participate in loss calculation.\\
        &\(c_f\) & The foreground regions that do not belong to target classes \(\mathcal{C}_t\). \\
        &\(c_u\) & Unknown classes defined in previous methods, consisting of \(\mathcal{C}_{1:t-1}\) and \(c'_u\).  \\
        &\(c_b\) & The background defined in previous methods, consisting of \(c'_b\) and \(c_u\).\\
        \midrule
        \multirow{8}{*}{\rotatebox{90}{Data and label}} 
        &\(\mathcal{D}_t\) & Training dataset of current incremental task $t$. \\
        &\(\mathcal{M}\) & Memory buffer, with fixed size of 100 for Pascal VOC and 300 for ADE20K. \\
        &\(x^t_i\) & The $i$-th image in \(\mathcal{D}_t\). \\
        &\(y^t_i\) & The pixel annotations of \(x^t_i\). \\
        &\(x^{m,t}_i\) & The mixed data selected from  \(\mathcal{M}\) and \(\mathcal{D}_t\). \\
        &\(\tilde{\mathcal{Y}}_{i}^{m,t}\) & The image-level pseudo-label of \(x^{m,t}_i\). \\
        &\(\tilde{y}_{i}^{p}\) & Label assigned to the permanent branch \(\phi_p\). \\
        &\(\tilde{y}_{i}^{t}\) & Label assigned to the temporary branch \(\phi_t\). \\
        \bottomrule
    \end{tabular}
    \label{tab:symbol}
\end{table}


\subsection{Observation}

\begin{figure}[h]
    \centering
    \includegraphics[width=0.75\linewidth]{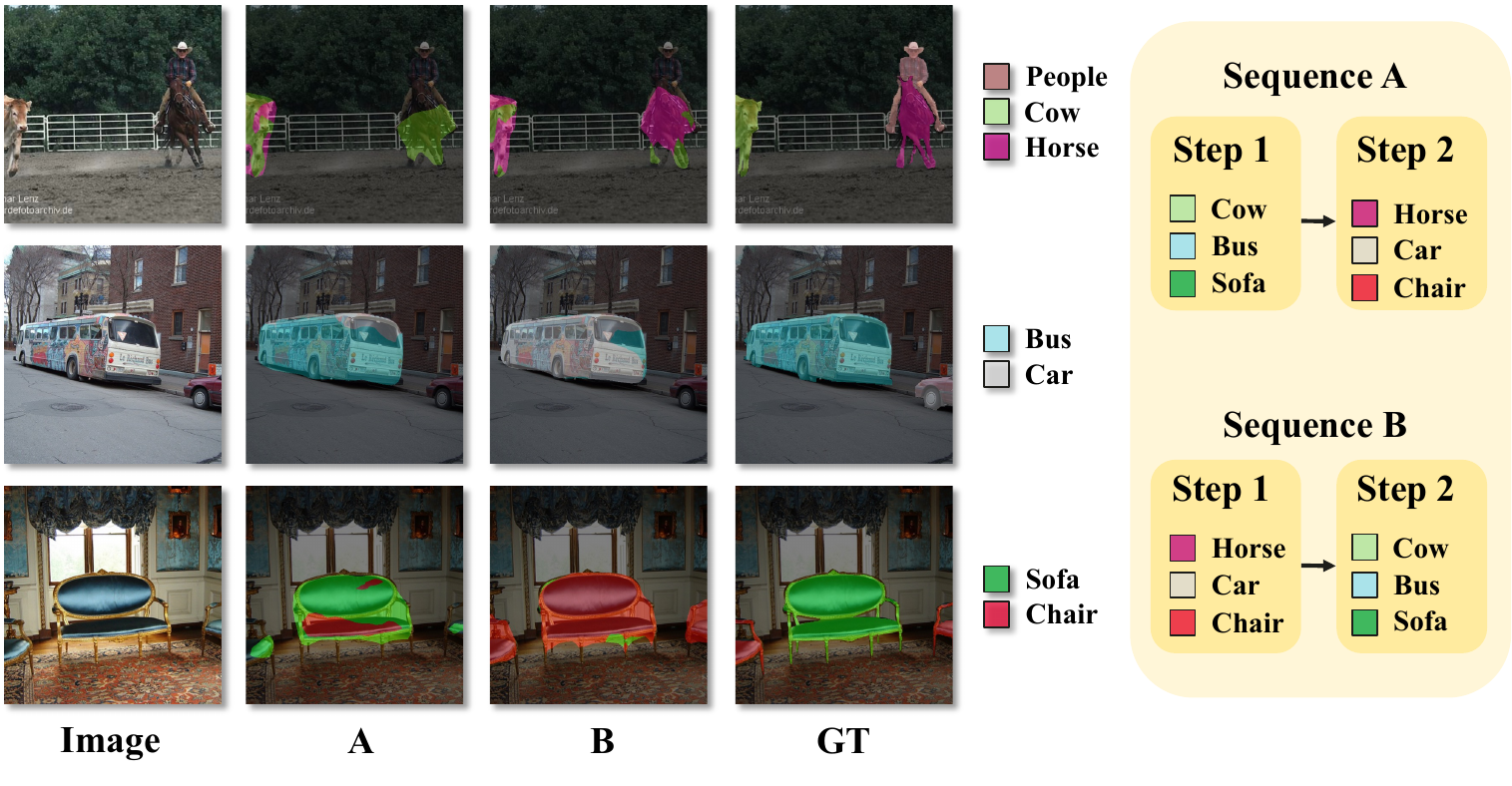}
    \caption{The visualization of separate optimization. \textit{Sequence A}: first learn ``cow'', ``bus'', ``sofa'' in step 1, then ``horse'', ``car'', ``chair'' in step 2. \textit{Sequence B}: first learn ``horse'', ``car'', ``chair'' in step 1, then ``cow'', ``bus'', ``sofa'' in step 2.
    }
    \label{fig:vis_so}
    \vspace{-5pt}
\end{figure}

\paragraph{Visualization of separate optimization}As shown in Figure~\ref{fig:vis_so}, to illustrate the inconsistent outputs caused by \textit{separate optimization}, we select three pairs of similar classes from Pascal VOC 2012: ``cow'' and ``horse'', ``bus'' and ``car'', ``sofa'' and ``chair'', and split them into two separate groups for learning. Sequence B follows a reverse learning order compared to Sequence A, and the goal is to examine the model's final predictions with different learning sequence. Columns A and B are the models' final predictions of sequence A and sequence B respectively. These visualizations indicate that the earlier head of the model tends to produce high scores for certain classes, regardless of the learning order, suggesting that \textit{separate optimization} causes a persistent bias towards the classes learned first.

\begin{figure}[t]
    \centering
    \includegraphics[width=1\linewidth]{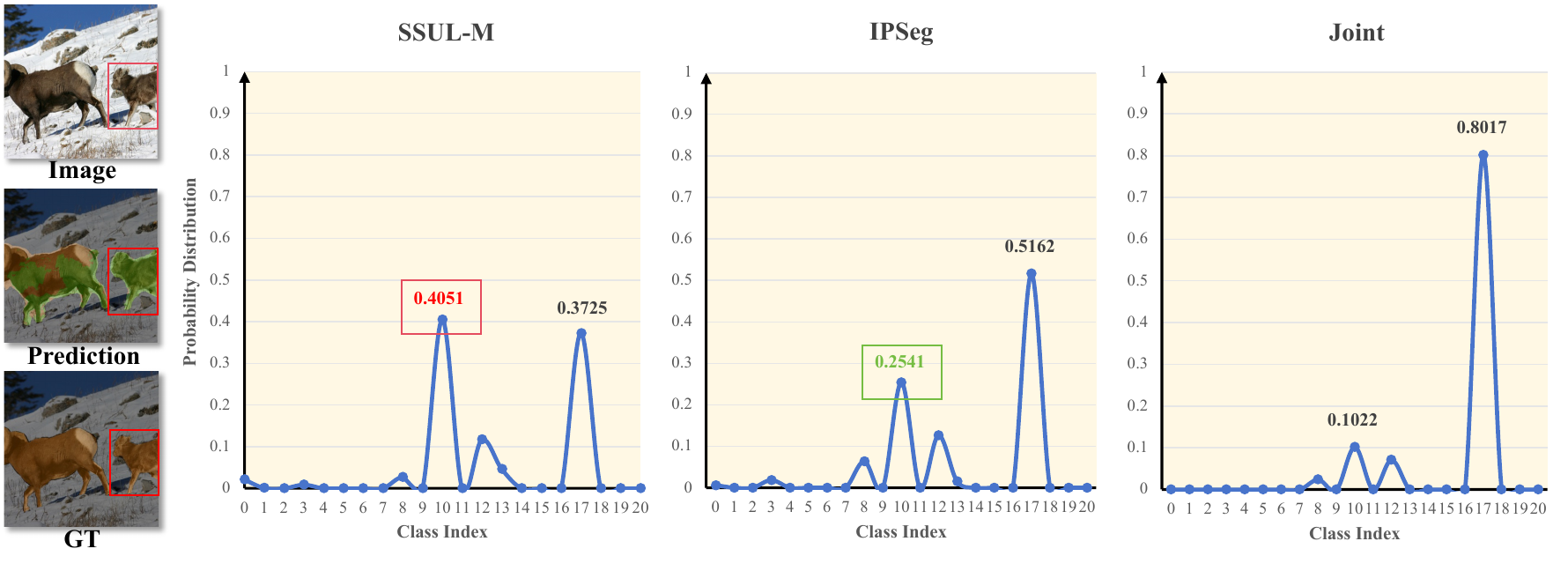}
    \caption{The probability distributions for SSUL-M, IPSeg, and Joint-Training (Joint) in the regions of incorrect predictions. Class indexes ``10'' and ``17'' represent ``cow'' and ``sheep'' respectively.}
    \label{fig:distribution}
\end{figure}

\paragraph{Impact of IPSeg on separate optimization}To validate the impact of IPSeg on \textit{separate optimization}, we calculate the average probability distribution of the incorrect prediction area (red box in the image) as depicted in Figure~\ref{fig:distribution}. SSUL-M misclassifies the little sheep as cow with abnormal probability distribution. In contrast,  IPSeg utilizes image posterior guidance to produce more accurate and harmonious prediction. Compared to previous works, IPseg maintains a harmonious and realistic probability distribution more similar to that of the theoretical upper bound, Joint-Training (Joint), demonstrating its superior capability in dealing with \textit{separate optimization}.

\begin{figure}[t]
    \centering
    \includegraphics[width=0.6\linewidth]{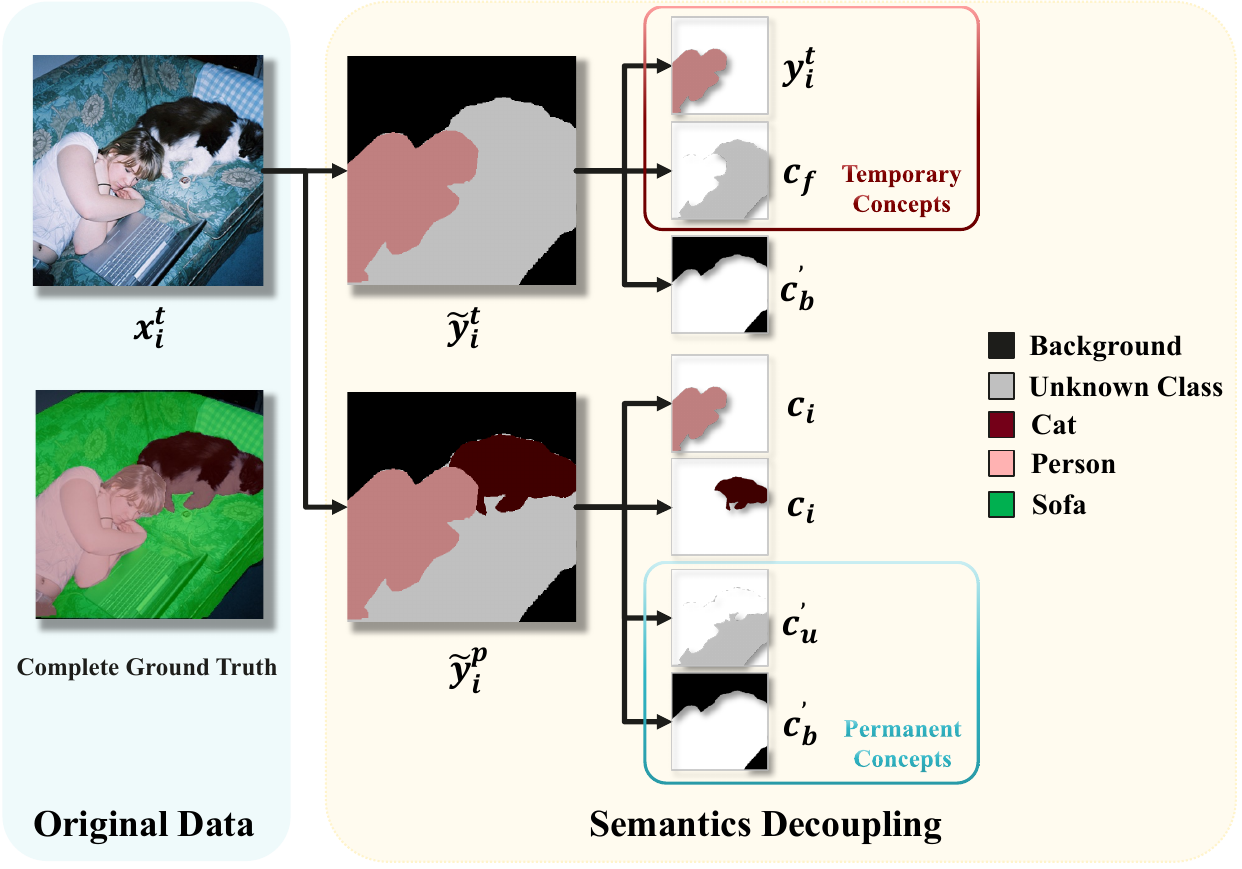}
    \caption{Semantics decoupling strategy}
    \label{fig:vis_KD}
    \vspace{-20pt}
\end{figure}

\paragraph{Visualization of semantics decoupling} Figure~\ref{fig:vis_KD} is the semantics decoupling illustration for image \(x^t_i\). In this case, the current classes \(\mathcal{C}_t\) ``person'' is provided with ground truth as \(y^t_i\). The foreground classes ``sofa'' and ``cat'', however, are unknown without ground truth. IPSeg uses a saliency map to locate the current unknown object ``sofa'' and ``cat'' as other foregrounds \(c_f\) and further utilizes pseudo label to distinguish ``sofa'' as unknown foreground \(c'_u\). It is worth noting that the regions of ``person'' and ``cat'' belong to the ignored regions \(c_i\) that do not participate in loss calculation. In this way, the remaining region is labeled as ``pure" background \(c'_b\). The ``pure" background \(c'_b\) and unknown foreground \(c'_u\) are considered as static and permanent concepts. The target classes \(\mathcal{C}_t\) with ground truth \(y^t_i\) and other foreground \(c_f\) are considered as dynamic and temporary concepts. 

\begin{figure}[H]
    \centering
    \includegraphics[width=0.9\linewidth]{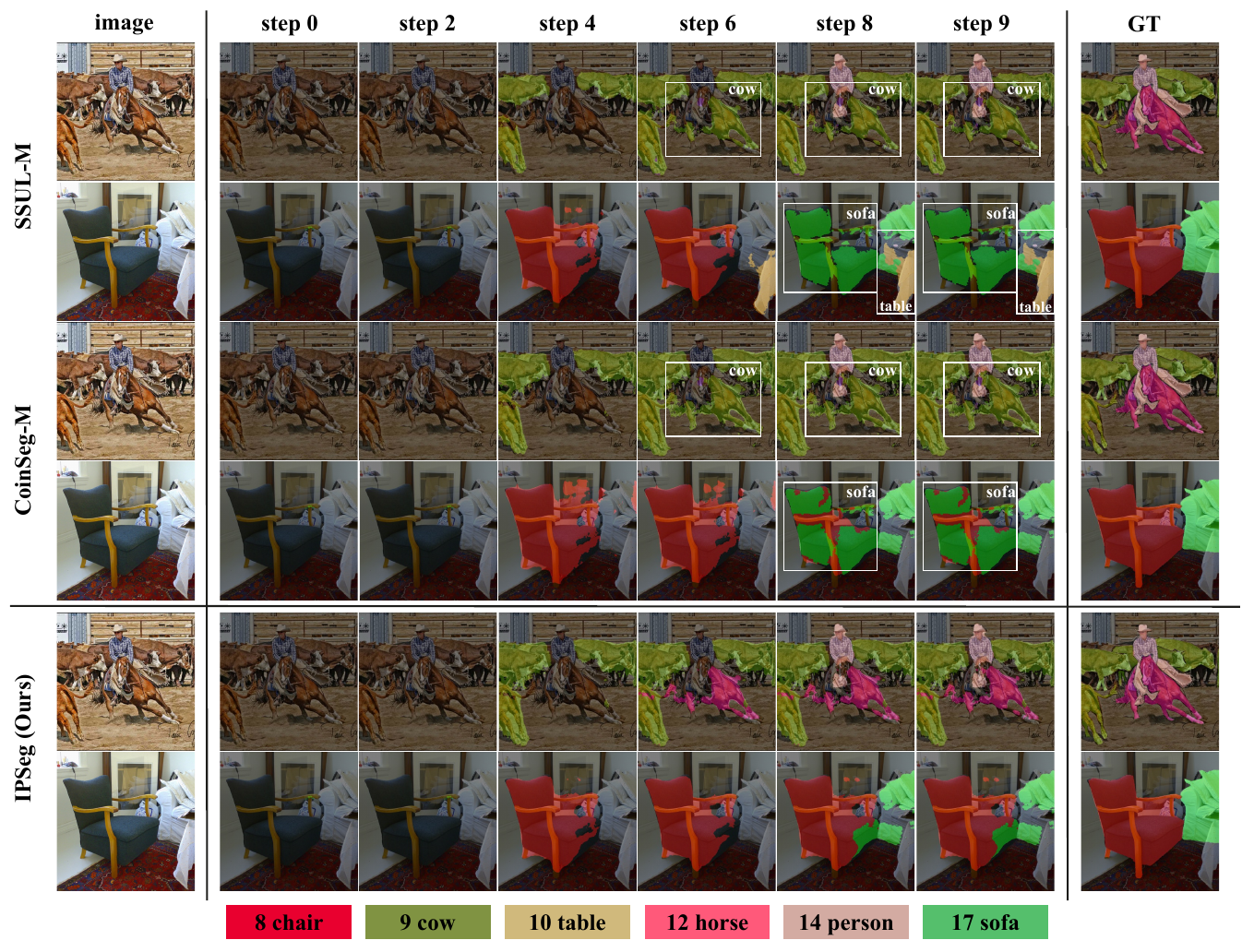}
    \caption{Qualitative analysis of IPSeg on Pascal VOC 2012. Texts and bounding boxes in white indicate incorrect class predictions starting from the corresponding incremental step. Color-shaded boxes with classes and indexes represent the learned classes in the corresponding learning steps.}
    \label{fig:vis_main}
    \vspace{-5pt}
\end{figure}

\paragraph{Qualitative analysis of IPSeg} Figure~\ref{fig:vis_main} presents a qualitative analysis of IPSeg compared with SSUL-M and CoinSeg-M. Visualization results are from each incremental step in the VOC 2-2 scenario. The results in rows 1, 3, and 5 demonstrate that both SSUL-M and CoinSeg-M mistakenly predict ``horse'' as ``cow'' at step 6, while IPSeg correctly identifies ``horse''.

\vspace{-10pt}
In rows 2, 4, and 6, IPSeg consistently predicts the old class ``chair'', whereas SSUL-M predicts ``sofa'' as ``table'' at step 6, and both SSUL-M and CoinSeg-M mistake ``chair'' for ``sofa'' at step 8. These visualization results reveal that IPSeg not only achieves excellent learning plasticity but also maintains strong memory stability.


\subsection{Additional Results and Analysis}

\paragraph{Evaluation on image-level predictions} To investigate the ability to resist catastrophic forgetting of the image posterior (IP) branch and the segmentation branch, we evaluate the image level accuracy performance of the base 15 classes using the IP branch and the segmentation branch at each step on Pascal VOC 15-1 as shown in Table~\ref{tab:IP_base_acc}. ``IP" refers to the image-level accuracy of the IP branch, ``Pixel" refers to the image-level accuracy of the segmentation branch, where class $\mathcal{C}$ exists if a pixel is predicted as $\mathcal{C}$. ``Pixel+IP" denotes the final result of IPSeg. The ablation shows that: the image-level performance suffers less forgetting than the segmentation, and our method shows similar property against forgetting with the help of IP.

\begin{table}[h]
    \centering
    \caption{The image-level accuracy of IP branch and the segmentation branch on 15 base classes.}
    \begin{tabular}{l|c c c c c c}
         \toprule
         ACC (\%) & step 0 & step 1  & step 2 & step 3 & step 4 & step 5 \\
         \midrule
         IP & 87.44 & 86.41 & 86.99 & 86.82 & 86.86 & 86.29 \\
         Pixel & 88.17 & 86.42 & 86.30 & 85.43 & 84.84 & 84.70 \\
         Pixel+IP & 93.07 & 92.24 & 92.41 & 91.93 & 91.95 & 91.02 \\
         \bottomrule
    \end{tabular}
    \label{tab:IP_base_acc}
\end{table}

We also present the image-level accuracy on all seen classes at each step to analyze their performance on both old classes and new classes in Table~\ref{tab:IP_all_acc}. For the segmentation branch, the image-level accuracy of it on all seen classes gradually degrades after learning new classes, performing worse than its accuracy on base classes. This indicates the segmentation branch performs poorly on new classes, which is consistent with our description about separate optimization. In contrast, the IP branch experiences less deterioration from separate optimization and help our method maintain a good balance between retaining old knowledge and learning new knowledge. 

\begin{table}[h]
    \centering
    \caption{The image-level accuracy of IP branch and the segmentation branch on all seen classes.}
    \begin{tabular}{l|c c c c c c}
         \toprule
         ACC (\%) & step 0 & step 1  & step 2 & step 3 & step 4 & \textbf{step 5 (Final)} \\
         \midrule
         IP & 87.44 & 82.54 & 81.14 & 81.32 & 82.09 & \textbf{82.34} \\
         Pixel & 88.17 & 83.56 & 82.29 & 78.23 & 77.60 & \textbf{76.57} \\
         Pixel+IP & 93.07 & 90.05 & 90.13 & 87.30 & 87.68 & \textbf{88.03} \\
         \bottomrule
    \end{tabular}
    \label{tab:IP_all_acc}
\end{table}

\paragraph{Evaluation on model parameters, training and inference costs} We provide a comprehensive analysis of the model parameters, training, and inference costs as shown in Table~\ref{tab:para_costs}. We test and report the results of IPSeg, SSUL-M and CoinSeg-M with Swin-B on the VOC 15-1 setting. We set \textit{image\_size=512x512, epochs=50, and batch\_size=16} in training and \textit{image\_size=512x512} for inference test. All results are run on RTX 3090 GPU.
\begin{itemize}
    \item \textbf{Model Parameters:} Using the thop tool, we analyze and compare the trainable parameters for these methods. The sizes of increased parameters in them are close, with average $3.84M$ per step. Additionaly, IPSeg has $29.72M$ parameters more than SSUL due to the additional image posterior branch.
    \item \textbf{Training:} Due to the introduced image posterior branch, IPSeg needs more training cost compared with SSUL-M but  less than CoinSeg-M. 
    \item \textbf{Inference:} The inference speed of IPSeg ($27.3$ FPS) is slightly lower than SSUL-M ($33.7$ FPS) and similar to CoinSeg-M ($28.2$ FPS). Due to the proposed image posterior branch, the model's floating-point operations ($137.1$ GFLOPs) are higher than the baseline ($94.9$ GFLOPs), and with an approximately $1$ GB increase in GPU usage. Note that the increase in FLOPs mainly stems from IPSeg’s use of image-level predictions to guide final outputs. Specifically, IPSeg broadcasts image-level predictions to match the shape of pixel-level logits and combines them through element-wise multiplication, which are inherently parallelizable and can be optimized and accelerated by GPUs, ensuring that the inference speed remains largely unaffected.
\end{itemize}

Overall, IPSeg introduces an additional image posterior branch with slight increases in model parameters, training cost, and inference but brings great performance improvement. It is a worthwhile trade-off between performance and cost.

\begin{table}[h]
    \centering
    \setlength{\tabcolsep}{3pt}  
    \caption{Comparison of IPSeg with baseline on model parameters, training and inference costs. }
    \resizebox{1.0\linewidth}{!}{
    \begin{tabular}{l||cccccc|cc|rcc}
        \toprule
         \multirow{2}{*}{Method} & \multicolumn{6}{c|}{Incremental Steps} & \multicolumn{2}{c|}{Training}& \multicolumn{3}{c}{Inference}\\
         & 0 & 1 & 2 & 3 & 4 & 5 & Time & GPU usage & FPS & FLOPs & GPU usage \\
         \midrule
         IPseg & 135.92 M & 139.76 M & 143.60 M & 147.66 M & 151.28 M & 155.12 M & 9h 14min & 21.1G & 27.3 & 137.1G & 6.2G \\
         SSUL-M & 106.20 M & 110.03 M & 113.89 M & 117.95 M & 121.56 M & 125.40 M & 7h 13min & 19.4G & 33.7 & 94.9G & 5.3G \\
         CoinSeg-M & 107.02 M & 111.15 M & 115.29 M & 119.42 M & 123.55 M & 127.68 M & $>$ 15h & 21.3G & 28.2 & 96.3G & 5.6G \\
         \bottomrule
    \end{tabular}
    }
    \label{tab:para_costs}
\end{table}

\paragraph{The details of efficient data storage in memory buffer}  

For raw data, IPSeg directly stores the image paths in a JSON file, as done in previous works~\citep{SSUL_cha2021ssul,microseg_zhang2022mining,coinseg_zhang2023coinseg}. For image-level labels, IPSeg stores the class labels of the images as arrays in the same JSON file with multi-hot encoding, where $1$ indicates the presence of a class and $0$ indicates absence. The memory cost for this is negligible. For pixel-level labels, instead of storing full-class annotations (with a data type of \textit{uint8} ) as prior approaches, IPSeg only stores the salient mask, where the background and foreground are labeled as $0$ and $1$, respectively (with a data type of \textit{bool} ). Theoretically, the storage space could be reduced to $1/8$.

\paragraph{Ablation study on memory buffer} Since IPSeg is designed for data-replay scenarios, the IP branch heavily relies on a memory buffer. To evaluate the impact of the memory buffer on performance, we compare the standard version of IPSeg with a data-free variant (denoted as IPSeg w/o M). As shown in Table~\ref{tab:impact_mem}, IPSeg demonstrates competitive performance even without the memory buffer. However, the performance gap between the data-free and data-replay settings highlights the essential role of the memory buffer in enhancing IPSeg's effectiveness.

\begin{table}[h]
    \centering
    \caption{Comparison with other methods in data-free version using Swin-B backbone.}
    \resizebox{0.75\linewidth}{!}{
    \begin{tabular}{l|ccc|ccc|ccc|ccc}
    \toprule
    \multirow{2}{*}{Method} & \multicolumn{3}{c|}{\textbf{VOC 15-5 (2 steps)}} &\multicolumn{3}{c|}{\textbf{VOC 15-1 (6 steps)}}  &\multicolumn{3}{c|}{\textbf{VOC 10-1 (11 steps)}} &\multicolumn{3}{c}{\textbf{VOC 2-2 (10 steps)}} \\
    & 0-15 & 16-20 & all & 0-15 & 16-20 & all & 0-10 & 11-20 & all & 0-2 & 3-20 & all \\
    \midrule
    SSUL & 79.7 & 55.3 & 73.9 & 78.1 & 33.4 & 67.5 & 74.3 & 51.0 & 63.2 & 60.3 & 40.6 & 44.0 \\
    MicroSeg & \textbf{81.9} & 54.0 & 75.2 & 80.5 & 40.8 & 71.0 & 73.5 & 53.0 & 63.8 & 64.8 & 43.4 & 46.5 \\
    IPSeg w/o M & 81.4 & \textbf{62.4} & \textbf{76.9} & \textbf{82.4} & \textbf{52.9} & \textbf{75.4} & \textbf{80.0} & \textbf{61.2} & \textbf{71.0} & \textbf{72.1} & \textbf{64.5} & \textbf{65.5} \\
    \midrule
    IPSeg w/ M & 83.3 & 73.3 & 80.9 & 83.5 & 75.1 & 81.5 & 80.3 & 76.7 & 78.6 & 73.1 & 72.3 & 72.4\\
    \bottomrule
    \end{tabular}
    }
    \label{tab:impact_mem}
    \vspace{-10pt}
\end{table}

\paragraph{Ablation study for hyper-parameters: weight terms of loss.} We conduct an ablation study on the two weight terms \(\lambda_1\) and \(\lambda_2\), testing values of 0.1, 0.25, 0.5, 0.75, and 1.0. The results are shown in Table~\ref{tab:loss_weight}. It is obvious that the setting of \(\lambda_1=0.5\) and \(\lambda_2=0.5\) achieves the best performance, which is the default value of IPSeg.

\paragraph{Ablation study of hyper-parameters} Table~\ref{tab:ablation_hyperpara}(a) and (b) illustrate the effects of hyper-parameters: memory size \(\left|\mathcal{M}\right|\), the strength of noise filtering \(\alpha_{\text{\tiny NF}}\), and background compensation \(\alpha_{\text{\tiny BC}}\), which shows that IPseg is not sensitive to the value of \(\alpha_{\text{\tiny NF}}\) and \(\alpha_{\text{\tiny BC}}\)
and we set the default values for these parameters to  \(\left|\mathcal{M}\right| = 100\), \(\alpha_{\text{\tiny NF}}=0.4\) and \(\alpha_{\text{\tiny BR}}=0.9\).

\begin{table}[h]
    \centering
    \caption{Ablation Studies on Pascal VOC 15-1 task for hyper-parameters: weight terms of loss,  \(\lambda_1\) and \(\lambda_2\).}
    \begin{tabular}{c|c c c c c}
         \toprule\(\)
         \(\lambda_1\) \textbackslash ~\(\lambda_2\) & 0.1 & 0.25 & 0.5 & 0.75 & 1.0 \\
         \midrule
         0.1 & 79.2 & 80.3 & 80.4 & 80.8 & 78.0 \\
         0.25 & 79.8 & 80.3 & 81.1 & 81.0 & 80.9 \\
         0.5 & 80.3 & 80.9 & \textbf{81.5} & 81.2 &  80.9 \\
         0.75 & 81.1 & 81.3 & 81.1 & 81.0 & 81.0 \\
         1.0 & 80.8 & 81.3 & 81.1 & 81.2 & 80.1 \\
         \bottomrule
    \end{tabular}
    \label{tab:loss_weight}
\end{table}

\begin{table}[t]
    \centering
    \caption{(a): Ablation studies for hyper-parameters: memory size \(|M|\), the ratio of noise filtering \(\alpha_{\text{\tiny NF}}\). (b): Ablation studies for hyper-parameters: the ratio of background compensation \(\alpha_{\text{\tiny BC}}\).}
    \begin{minipage}{0.5\textwidth}
        \centering
        \resizebox{0.8\textwidth}{!}{
            \begin{tabular}{c||c|ccc}
            \toprule
             & value  &  0-15 & 16-20 & all \\
            \midrule
            {\multirow{3}{*}{\(|M|\)}} & 50 & 83.4 & 72.3 & 80.8 \\
            & 100 & 83.5 & 75.1 & 81.5 \\
            & 200 & 83.5 & 75.5 & 81.7 \\
            \midrule
            {\multirow{5}{*}{\(\alpha_{\text{\tiny NF}}\)}}
            & 0.2 & 83.5 & 75.0 & 81.4 \\
            & 0.4 & \textbf{83.6} & \textbf{75.1} & \textbf{81.6} \\
            & 0.6 & 83.4 & 74.6 & 81.3 \\
            & 0.8 & 83.3 & 74.2 & 81.2 \\
            & 1.0 & 83.4 & 74.7 & 81.3 \\
            \bottomrule
            \end{tabular}
        }
        \caption*{(a)}
    \end{minipage}\hfill
    \begin{minipage}{0.5\textwidth}
        \centering
        \resizebox{0.66\textwidth}{!}{
            \begin{tabular}{c | ccc}
            \toprule
            {\multirow{2}{*}{ \(\alpha_{\text{\tiny BC}}\)}} & \multicolumn{3}{c}{\textbf{VOC 15-1 (6 steps)}} \\ & 0-15 & 16-20 & all \\
            \midrule
            1 & 82.9 & 74.9 & 81.0 \\
            0.9 & \textbf{83.5} & \textbf{75.1} & \textbf{81.6} \\
            0.8 & 83.5 & 75.0 & 81.5\\
            0.7 & 83.4 & 74.4 & 81.3\\
            0.6 & 83.3 & 74.3 & 81.2\\
            0.5 & 83.2 & 73.8 & 81.0\\
            0 & 80.6 & 66.6 & 77.3\\
            \bottomrule
            \end{tabular}
        }
        \caption*{(b)}
    \end{minipage}
    \label{tab:ablation_hyperpara}
\end{table}

\paragraph{Impact of semantics decoupling on temporary concepts} To understand which types of concepts most benefit from IPSeg, we categorize the 20 classes of Pascal VOC into 15 base classes and 5 new classes based on the incremental process. According to the learning objectives, all foreground classes are treated as temporary concepts in the corresponding step and the background is constantly considered as permanent ones.
The comparison shown in Table~\ref{tab:overall_results_of_KD} indicates that the new classes gain more significant performance improvement than the base classes. Furthermore, the permanent concepts (i.e., the background) achieve less improvement compared to the temporary concepts. This observation suggests that IPSeg is more effective in enhancing the learning of new foreground classes.

\begin{table}[h]
    \caption{The ablation study of \textbf{SD} over background (BG), base foreground classes (Base) and new foreground classes (New) on Pascal VOC 15-1 with Swin-B backbone.}
    \centering
    \resizebox{0.55\linewidth}{!}{
    \begin{tabular}{c || ccc|c}
    \toprule
    \textbf{SD} & BG & Base(1-15) & New(16-20) & All \\
    \midrule
    \XSolidBrush & 92.4 & 78.5 & 69.6 & 77.0 \\
    \Checkmark & 94.3(+1.9) & 82.3(+3.7) & 75.1(\textbf{+5.5}) & 81.5 \\
    \bottomrule
    \end{tabular}
    }
    \label{tab:overall_results_of_KD}
\end{table}

\paragraph{Detailed results of semantics decoupling on temporary concepts} To understand which types of classes or concepts benefit from our method, we compare IPseg against the baseline on 20 classes of the VOC 15-1 setting. The detailed results are presented in Table~\ref{tab:detailed_results_of_KD}, which demonstrates that IPSeg is more effective in enhancing the learning of new foreground classes.

\begin{table}[h]
    \centering
    \caption{Detailed results of the ablation study for semantics decoupling (\textbf{SD}) over each class on Pascal VOC 15-1 with Swin-B backbone. Texts in red indicate 5 new classes.}
    \begin{tabular}{c||c|c|c|c|c|c|c|c|c|c|c}
    \toprule
    \textbf{SD} & \multicolumn{11}{c}{Detailed results} \\
    \midrule
    \multirow{4}{*}{\XSolidBrush}
    & BG & plane & bike & bird & boat & bottle & bus & car & cat & chair & cow \\ 
    & 92.4 & 87.4 & 37.7 & 89.1 & 67.8 & 80.4 & 93.8 & 86.9 & 93.6 & 43.9 & 85.7 \\
    & tabel & dog & horse & motor & person & \textcolor{red}{plant} & \textcolor{red}{sheep} & \textcolor{red}{sofa} & \textcolor{red}{train} & \textcolor{red}{TV} & \textbf{mIoU}  \\
    & 63.8 & 90.0 & 87.2 & 85.9 & 84.1 & \textcolor{red}{57.5} & \textcolor{red}{81.6} & \textcolor{red}{53.3} & \textcolor{red}{87.1} & \textcolor{red}{68.4} & \textbf{77.0}\\
    
    \midrule
    \multirow{4}{*}{\Checkmark}
    & BG & plane & bike & bird & boat & bottle & bus & car & cat & chair & cow \\ 
    & 94.3 & 91.8 & 43.8 & 93.8 & 75.0 & 86.0 & 94.2 & 91.2 & 96.1 & 44.3 & 94.6\\
    & tabel & dog & horse & motor & person & \textcolor{red}{plant} & \textcolor{red}{sheep} & \textcolor{red}{sofa} & \textcolor{red}{train} & \textcolor{red}{TV} & \textbf{mIoU}  \\
    & 67.3 & 94.5 & 93.0 & 88.8 & 88.2 & \textcolor{red}{65.6} & \textcolor{red}{90.1} & \textcolor{red}{57.9} & \textcolor{red}{89.3} & \textcolor{red}{72.7}  & \textbf{81.5}\\
    \bottomrule
    \end{tabular}
    \label{tab:detailed_results_of_KD}
\end{table} 

\paragraph{Class-wise results of IPSeg}Table~\ref{tab:detailed_results} shows the detailed experimental results of IPSeg for each class across four incremental scenarios of Pascal VOC 2012. IPSeg demonstrates superior performance in various incremental learning tasks, including standard tasks with a large number of initial classes (e.g., 15-5 and 15-1) and long-range tasks with fewer initial classes (e.g., 10-1 and 2-2). Notably, in the 2-2 task, the mIoU for ``cow" and ``horse" reaches $70.7$\% and $81.4$\%, respectively. This indicates that IPSeg maintains excellent prediction performance even when classes with similar semantic information are trained in different stages.

\begin{table}[h]
    \caption{Class-wise results of IPSeg over each class.}
    \centering
    \resizebox{0.8\linewidth}{!}{
    \begin{tabular}{l||c|c|c|c|c|c|c|c|c|c|c}
    \toprule
    \multirow{4}{*}{\textbf{VOC 15-5}}
    & BG & plane & bike & bird & boat & bottle & bus & car & cat & chair & cow \\ 
    & 93.6 & 92.2 & 44.9 & 93.8 & 74.4 & 85.2 & 93.9 & 90.8 & 96.2 & 43.0 & 94.5 \\
    & tabel & dog & horse & motor & person & plant & sheep & sofa & train & TV & \textbf{mIoU}  \\
    & 68.0 & 94.2 & 92.9 & 88.0 & 88.0 & 66.3 & 91.5 & 46.4 & 87.8 & 74.4  & \textbf{80.9}\\
    \midrule
    \multirow{4}{*}{\textbf{VOC 15-1}}
    & BG & plane & bike & bird & boat & bottle & bus & car & cat & chair & cow \\ 
    & 94.3 & 91.8 & 43.8 & 93.8 & 75.0 & 86.0 & 94.2 & 91.2 & 96.1 & 44.4 & 93.5 \\
    & tabel & dog & horse & motor & person & plant & sheep & sofa & train & TV & \textbf{mIoU}  \\
    & 67.4 & 94.5 & 92.9 & 88.8 & 88.2 & 66.4 & 88.6 & 58.1 & 89.5 & 72.7 & \textbf{81.5}\\
    \midrule
    \multirow{4}{*}{\textbf{VOC 10-1}}
    & BG & plane & bike & bird & boat & bottle & bus & car & cat & chair & cow \\ 
    & 93.1 & 93.1 & 42.2 & 93.2 & 72.1 & 83.5 & 93.9 & 91.9 & 95.8 & 38.0 & 86.9 \\
    & tabel & dog & horse & motor & person & plant & sheep & sofa & train & TV & \textbf{mIoU}  \\
    & 54.1 & 91.8 & 86.1 & 87.5 & 87.5 & 64.2 & 85.6 & 49.9 & 88.4 & 72.1 & \textbf{78.6}\\
    \midrule
    \multirow{4}{*}{\textbf{VOC 2-2}}
    & BG & plane & bike & bird & boat & bottle & bus & car & cat & chair & cow \\ 
    & 91.4 & 88.6 & 39.3 & 87.9 & 71.5 & 71.9 & 89.1 & 78.2 & 89.3 & 28.8 & 70.7\\
    & tabel & dog & horse & motor & person & plant & sheep & sofa & train & TV & \textbf{mIoU}  \\
    & 55.2 & 83.5 & 84.1 & 77.6 & 82.6 & 63.6 & 72.1 & 44.4 & 82.1 & 69.1 & \textbf{72.4}\\
    \bottomrule
    \end{tabular}
    }
    \label{tab:detailed_results}
\end{table}

\paragraph{Experimental results of disjoint setting}
To demonstrate IPSeg's robust learning capability under different incremental learning settings and to further prove its superiority over general methods, we evaluate IPSeg using the disjoint setting on the Pascal VOC 2012 dataset for the 15-1 and 15-5 tasks, as shown in Table~\ref{tab:voc_res_disjoint}. The results indicate that IPSeg consistently achieves the best performance compared to state-of-the-art methods. Additionally, similar to the results in the overlap setting, IPSeg exhibits a strong ability to learn new classes while retaining knowledge of the old classes. Specifically, IPSeg outperformed the second-best method by $\textbf{10.1}$\% in the 15-5 task and by $\textbf{22.8}$\% in the 15-1 task in terms of new class performance.

\begin{table}[h]
    \caption{Comparison with state-of-the-art methods on Pascal VOC 2012 dataset for disjoint setup.}
    \centering
    \resizebox{0.53\linewidth}{!}{
    \begin{tabular}{c|c||ccc|ccc}
    \toprule
    \multicolumn{2}{c |}{\multirow{2}{*}{Method}}  & \multicolumn{3}{c|}{\textbf{VOC 15-5 (2 steps)}} &\multicolumn{3}{c}{\textbf{VOC 15-1 (6 steps)}}  \\
     \multicolumn{2}{c |}{} & 0-15 & 16-20 & all & 0-15 & 16-20 & all \\
    \midrule
    \multirow{4}{*}{\rotatebox{90}{Data-free}}
    & LwF-MC & 67.2 & 41.2 & 60.7 & 4.5 & 7.0 & 5.2\\
    & ILT & 63.2 & 39.5 & 57.3 & 3.7 & 5.7 & 4.2\\
    & MiB & 71.8 & 43.3 & 64.7 & 46.2 & 12.9 & 37.9\\
    & RCIL & 75.0 & 42.8 & 67.3 & 66.1 & 18.2 & 54.7\\
    \midrule
    \multirow{5}{*}{\rotatebox{90}{Replay-based}} 
    & SDR & 74.6 & 44.1 & 67.3 & 59.4 & 14.3 & 48.7\\
    & SSUL-M & 76.5 & 48.6 & 69.8 & 76.5 & 43.4 & 68.6\\
    & MicroSeg-M & 80.7 & 55.2 & 74.7 & 80.0 & 47.6 & 72.3\\
    & CoinSeg-M & 82.9 & 61.7 & 77.9 & 82.0 & 49.6 & 74.3\\
    & IPSeg(ours) & \textbf{82.7} & \textbf{71.8} & \textbf{80.1} & \textbf{82.6} & \textbf{72.4} & \textbf{80.2}\\
    \bottomrule
    \end{tabular}
    }
    \label{tab:voc_res_disjoint}
\end{table}

\begin{figure}[H]
    \centering
    \includegraphics[width=1\linewidth]{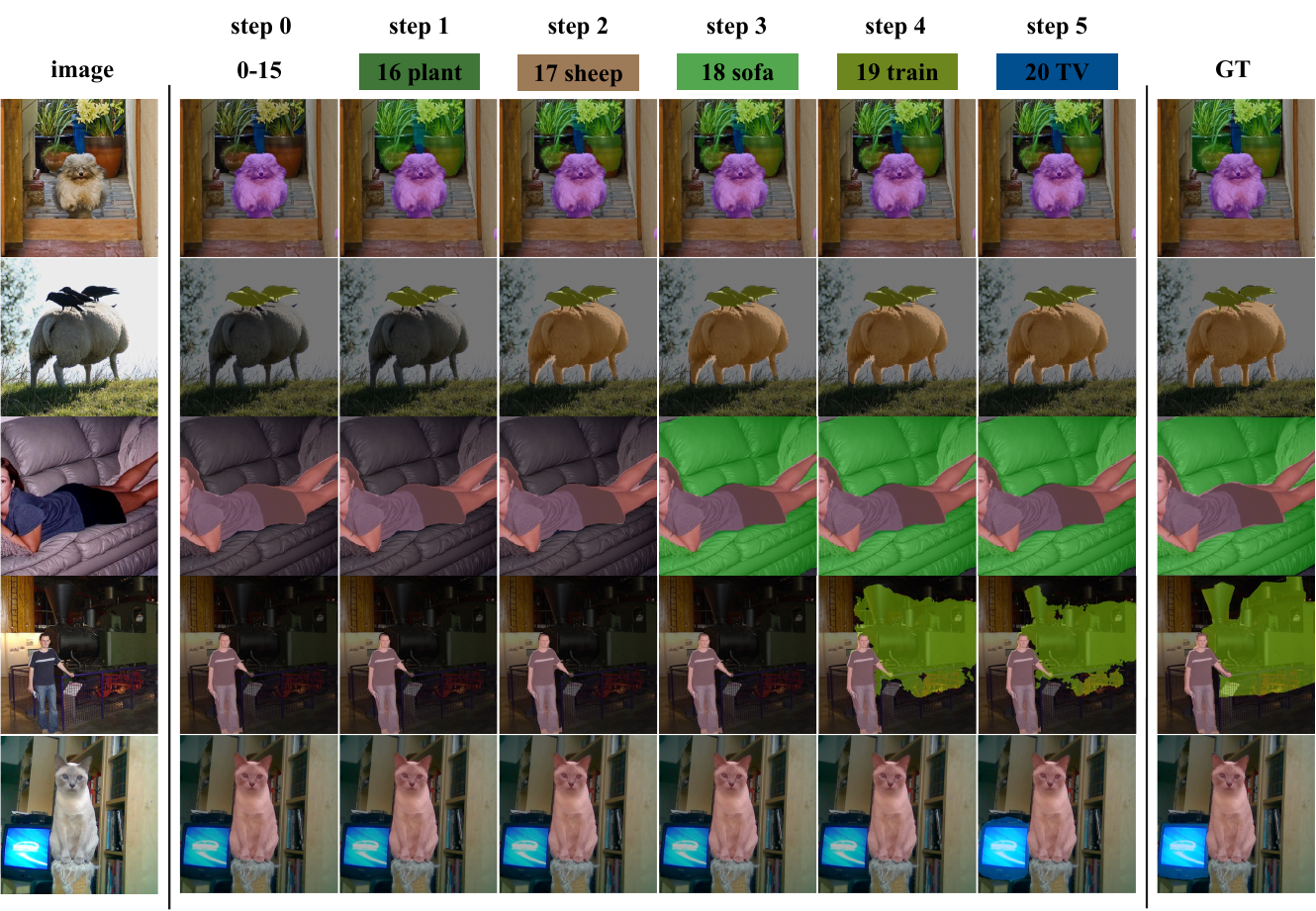}
    \caption{Qualitative results on Pascal VOC 2012 dataset with the 15-1 scenario.}
    \label{fig:vis_voc}
\end{figure}

\paragraph{More qualitative results on Pascal VOC 2012}
In addition to the qualitative results of the VOC 2-2 task shown in the main paper, we present additional qualitative analysis in Figure~\ref{fig:vis_voc}. We select the 15-1 task and perform a visual analysis for each newly added class. Each image includes both old and new classes, covering indoor and outdoor scenes as well as various objects and environments. For example, the first row shows the learning ability for "plant". After step 1, the model consistently predicts ``plant" correctly while retaining the ability to recognize "dog". Similarly, rows 2-5 show consistent performance. For each new class (i.e., sheep, sofa, train, and TV in the figure), IPSeg quickly adapts to them while retaining the ability to recognize old classes (i.e., bird, person, cat in the figure). These results clearly and intuitively demonstrate IPSeg's strong capability in addressing incremental tasks.

\begin{figure}[H]
    \centering
    \includegraphics[width=1\linewidth]{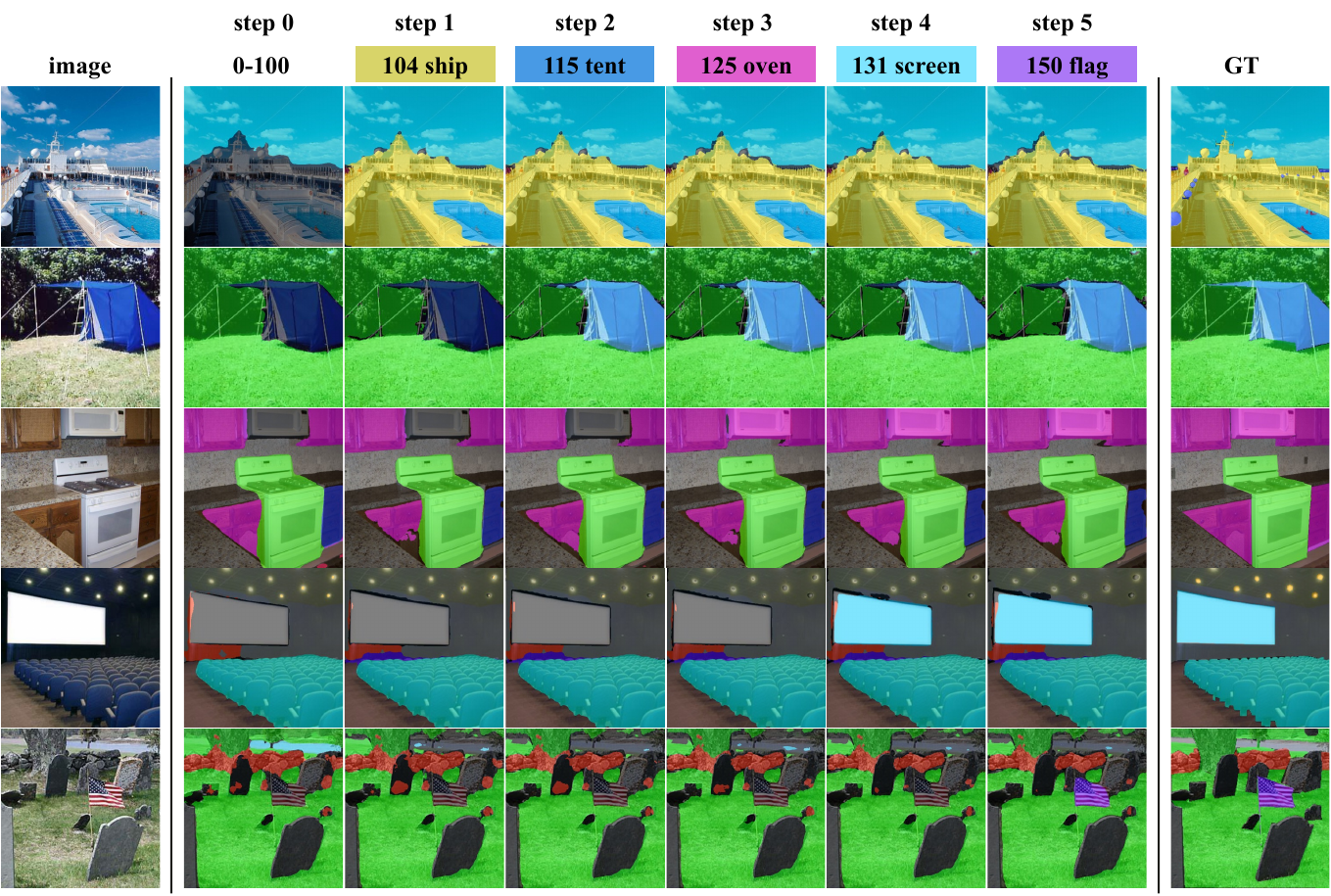}
    \caption{Qualitative results on ADE20K dataset with the 100-10 scenario.}
    \label{fig:vis_ade}
\end{figure}

\paragraph{More qualitative results on ADE20K}
The qualitative results of the 100-10 task on the ADE20K dataset are shown in Figure~\ref{fig:vis_ade}. We select five examples to illustrate the model's ability to predict various classes as the learning step increases. Row 1 shows the performance of predicting the new class "ship" in step 1, where the model effectively recognizes both the old class "sky" and the new class "ship." Similarly, in rows 2-5, for the newly introduced classes (tent, oven, screen, flag), IPSeg demonstrates excellent performance in learning the new classes without forgetting the old ones. This indicates that IPSeg achieves a balance between stability and plasticity even on more challenging 
 and realistic datasets.

\newpage

\end{document}